\documentclass{article}
\usepackage{microtype}
\usepackage{graphicx}
\usepackage{caption}
\usepackage{subcaption}
\usepackage{booktabs} 

\usepackage{hyperref}

\usepackage{fancyhdr}
\usepackage[utf8]{inputenc} 
\usepackage{url}            
\usepackage{amsfonts}       
\usepackage{nicefrac}       
\usepackage{amsmath}
\usepackage{amssymb}
\usepackage{commath}
\usepackage{mathtools}
\usepackage{multirow}
\usepackage{dsfont}
\usepackage{xspace}
\usepackage[ruled,vlined]{algorithm2e}
\usepackage{algorithm}

\usepackage[accepted]{icml2021}

\usepackage{amsmath,amsfonts,bm}

\def\Secref#1{Section~\ref{#1}}

\def\eqref#1{equation~\ref{#1}}
\def\Eqref#1{Equation~\ref{#1}}

\def\1{\bm{1}}

\DeclareMathAlphabet{\mathsfit}{\encodingdefault}{\sfdefault}{m}{sl}
\SetMathAlphabet{\mathsfit}{bold}{\encodingdefault}{\sfdefault}{bx}{n}

\def\gO{{\mathcal{O}}}

\def\gT{{\mathcal{T}}}

\def\sA{{\mathbb{A}}}

\def\sD{{\mathbb{D}}}

\def\sI{{\mathbb{I}}}

\def\sM{{\mathbb{M}}}

\def\sO{{\mathbb{O}}}

\def\sS{{\mathbb{S}}}

\DeclareMathOperator*{\argmax}{arg\,max}
\DeclareMathOperator*{\argmin}{arg\,min}

\newtheorem{assumption}{Assumption}

\newtheorem{theorem}{Theorem}
\newtheorem{definition}{Definition}
\newtheorem{lemma}{Lemma}

\newcommand{\method}{Option-GAIL\xspace}
\newcommand{\email}[1]{\textless #1\textgreater}

\begin{document}

\twocolumn[
\icmltitle{Adversarial Option-Aware Hierarchical Imitation Learning}



\icmlsetsymbol{equal}{*}

\begin{icmlauthorlist}
	\icmlauthor{Mingxuan Jing}{tsinghua}
	\icmlauthor{Wenbing Huang}{tsinghua}
	\icmlauthor{Fuchun Sun$^\dag$}{tsinghua,bosch}
	\icmlauthor{Xiaojian Ma}{ucla}
	\icmlauthor{Tao Kong}{bytd}
	\icmlauthor{Chuang Gan}{mit}
	\icmlauthor{Lei Li}{bytd}
\end{icmlauthorlist}

\icmlaffiliation{tsinghua}{Department of Computer Science and Technology, Tsinghua University, Beijing, China (Mingxuan Jing \email{jingmingxuan@outlook.com}; Wenbing Huang \email{hwenbing@126.com}; Fuchun Sun \email{fcsun@tsinghua.edu.cn})}
\icmlaffiliation{ucla}{University of California, Los Angeles, USA}
\icmlaffiliation{bytd}{Bytedance AI Lab, Beijing, China}
\icmlaffiliation{mit}{MIT-IBM Watson AI Lab, USA}
\icmlaffiliation{bosch}{THU-Bosch JCML center}

\icmlcorrespondingauthor{$^\dag$Fuchun Sun}{fcsun@mail.tsinghua.edu.cn}

\icmlkeywords{Options, Imitation Learning, Hierarchical}

\vskip 0.3in
]



\printAffiliationsAndNotice{}  

\begin{abstract}
It has been a challenge to learning skills for an agent from long-horizon unannotated demonstrations. Existing approaches like Hierarchical Imitation Learning(HIL) are prone to compounding errors or suboptimal solutions. 
In this paper, we propose \method, a novel method to learn skills at long horizon. 
The key idea of \method is modeling the task hierarchy by options and train the policy via generative adversarial optimization. In particular, we propose an Expectation-Maximization(EM)-style algorithm: an E-step that samples the options of expert conditioned on the current learned policy, and an M-step that updates the low- and high-level policies of agent simultaneously to minimize the newly proposed option-occupancy measurement between the expert and the agent.
We theoretically prove the convergence of the proposed algorithm.
Experiments show that \method outperforms other counterparts consistently across a variety of tasks. 

\end{abstract}

\section{Introduction}
\label{sec:intro}
Hierarchical Imitation Learning (HIL) has exhibited promises on acquiring long-term skills directly from demonstrations~\citep{h-bc,hil2,hil3,hil4,hil5}. It contends the nature of sub-task decomposition in long-horizontal tasks, enjoying richer expressivity on characterizing complex decision-making than canonical Imitation Learning (IL). In general, HIL designs micro-policies for accomplishing the specific control for each sub-task, and employs a macro-policy for scheduling the switching of micro-policies. Such a two-level decision-making process is usually formulated by an Option model~\citep{SMDP} or goal-based framework~\citep{h-bc}. Although some works~\citep{tripleGAIL} have assumed the help of sub-task segmentation annotations, this paper mainly focuses on learning from unsegmented demonstrations to allow more practicability.

Plenty of HIL methods have been proposed, including the Behavior-Cloning(BC) based and the Inverse Reinforcement Learning(IRL) based approaches. For the first avenue, \citet{provable_h-bc, h-bc} customize BC to hierarchical modeling~\citep{ML16}, which, unfortunately, is prone to compounding errors~\citep{compounding-error} in case that the demonstrations are limited in practice. On the contrary, IRL is potential to avoid compounding errors by agent self-explorations. 
However, addressing IRL upon the Option model, by no means, is non-trivial considering the temporal coupling of the high-level and low-level policies. The works by \citet{d-info-gail,d-info-gail2} relax this challenge by training the two-level policies separately. Nevertheless, this two-stage method potentially leads to sub-optimal solutions in the absence of training collaboration between the two stages.

To overcome the aforementioned issues, this work proposes a novel Hierarchical IRL framework---\method, which is theoretically elegant, free of compounding errors, and end-to-end trainable. Basically, it is built upon GAIL~\citep{GAIL}  with two nutritive enhancements: 1) replacing Markov Decision Process (MDP) with the Option model; 2) augmenting the occupancy measurement matching with options. Note that GAIL is a popular IL method that casts policy learning upon MDP into an occupancy measurement matching problem. Therefore, it is natural to replace the MDP with the Option model for hierarchical modeling. Besides, GAIL mimicking a policy from an expert is equivalent to matching its occupancy measurement. This equivalence holds when the policy is one-to-one corresponding to its induced occupancy measurement, which, yet, is not valid in our hierarchical context. As we will depict in the paper, the policy of HIL is related to options, and its one-to-one correspondence only exists with regard to option-extended occupancy measurement other than traditional occupancy measurement without options. Hence, it is indispensable to involve options into the matching goal in our second enhancement.  
Notably, the option switching is inherently guaranteed in our model, without extra regulators such as mutual information used in~\citep{d-info-gail} to encourage the correlation between action-state pairs and options. 

It is non-straightforward to train our \method, specifically when the expert options are unavailable. We thereby propose an EM-like learning algorithm to remedy the training difficulty. Specifically, in the E-step, we apply a Viterbi method~\citep{viterbi} to infer the expert options conditional on the agent policy and expert actions/states. For the M-step, with the expert options provided, we optimize both the high- and low-level policies to minimize the discrepancy between the option-occupancy measurements of expert and agent. To be more specific, the M-step is implemented by a min-max game: maximizing a discriminator to estimate the discrepancy, and updating the policy to minimize the discriminative cost via a hierarchical RL method~\citep{DAC}. Notably, the convergence of the proposed EM algorithm is theoretically guaranteed if certain mild conditions hold. 

In summary, our main contributions include:
\begin{itemize}
    \item We propose \method, a theoretically-grounded framework that integrates hierarchical modeling with option-occupancy measurement matching. It is proved that \method ensures the discrepancy minimization between the policies of demonstrator and imitator.  

    \item We propose an EM-like algorithm to train the parameters of \method end-to-end. This method alternates option inference and policy updating and is proved to converge eventually.
    
    \item We evaluate our proposed method on several robotic locomotion and manipulation tasks against state-of-the-art HIL/IL baselines. The results demonstrate that our approach attains both dramatically faster convergence and better final performance over the counterparts. A complete set of ablation studies also verify the validity of each component we proposed. 
\end{itemize}

\section{Related Works}
\label{sec:related}
There have already been plenty of works researching on HIL, which, in general, can be categorized into two classes: Hierarchical Behavior Cloning (H-BC) and Hierarchical Inverse Reinforcement Learning (H-IRL).

\textbf{Hierarchical BC.}~~In these avenues, HIL is an extension of Behavior Cloning (BC)~\citep{bc1,bc2}. As a result, the missing sub-task information needs to be provided or inferred to ensure the learnability of hierarchical policies. For example, \citet{h-bc} require predefined sub-goals when learning the goal-based policies; \citet{Compile} try to alleviate the requirement of sub-task information by formulating the policy learning as an unsupervised trajectory soft-segmentation problem; \citet{ML16} and \citet{provable_h-bc} employ Baum-Welch algorithm~\citep{BW-alg} upon the option framework to infer the latent option from demonstrations, and directly optimize both the high- and low-level policies. Despite its easy implementation, behavior cloning is vulnerable to compounding errors~\citep{compounding-error} in case of the limited demonstrations, while our method enjoys better sample efficiency thanks to the IRL formulation.

\begin{figure} 
    \begin{center}
        \includegraphics[width=0.8\linewidth]{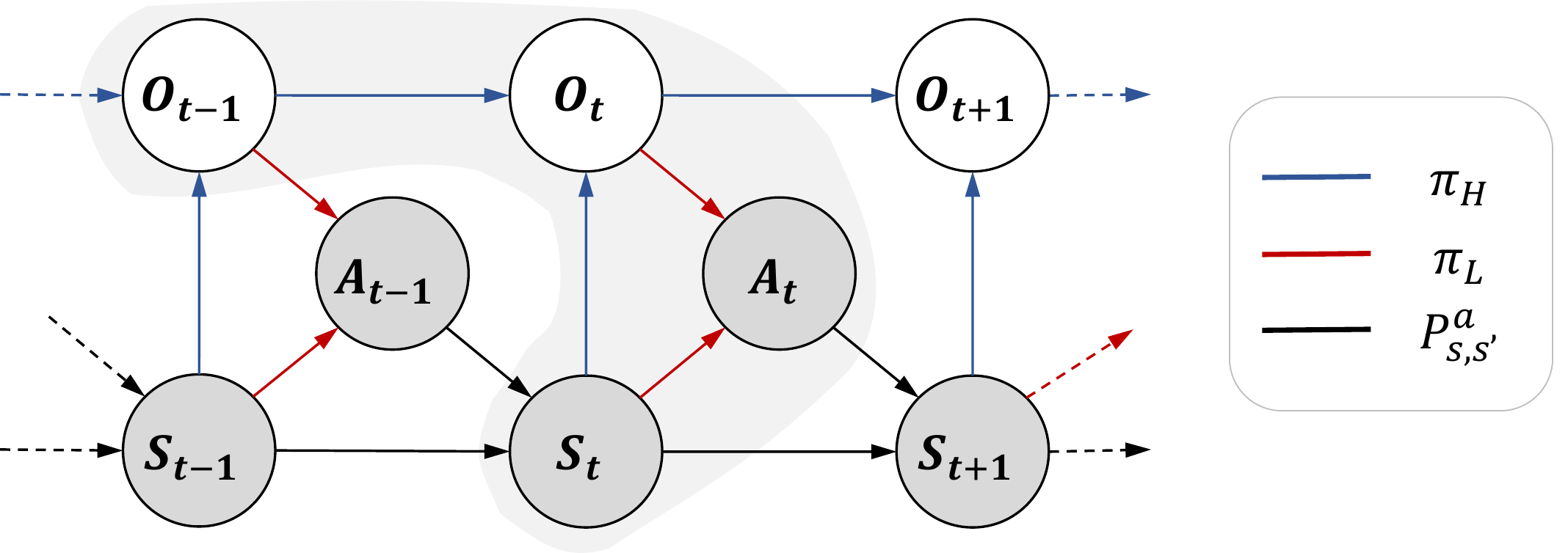} 
        \caption{\label{fig:one-step-opt-PGM} The probabilistic graph of the one-step option model. The shade masks the group of nodes that induce our option-occupancy measurement.}
    \end{center} \end{figure}

\textbf{Hierarchical IRL.}~~Contrasted to the BC-based approaches, Hierarchical IRL avoids compounding errors by taking advantage of the agent's self-exploration and reward reconstruction.
Following the Generative Adversarial Imitation Learning (GAIL)~\citep{GAIL}, the works by~\citep{d-info-gail,d-info-gail2} realize HIL by introducing a regularizer into the original IRL problem and maximizing the directed information between generated trajectories and options. However, the high- and low-level policies are trained separately in these approaches: the high-level policy is learned with behavior cloning and remains fixed during the GAIL-based low-level policy learning. As we reported in the experiments, such a two-staged paradigm
could lead to potentially poor convergence compared to our end-to-end training fashion.
\citet{optionGAN} propose an end-to-end HIL method without pre-training. However, it adopts a Mixture-of-Expert (MoE) strategy rather than the canonical option framework. Therefore, an option is exclusively determined by the corresponding state, ignoring its relation to the option of the previous step (Figure~\ref{fig:one-step-opt-PGM}). On the contrary, our method conducts option inference and policy optimization that are strictly amenable to the option dynamics~\citep{SMDP}, thus delineating the hierarchy of sub-tasks in a more holistic manner. 

\section{Preliminaries}
\label{sec:prelim}
This section briefly introduces preliminary knowledge and notation used in our work.

\textbf{Markov Decision Process:}\label{mdp}
A Markov Decision Process (MDP) is a 6-element-tuple $\left(\sS, \sA, P_{s,s'}^a, R_s^a, \mu_0, \gamma\right)$, where $\left(\sS,\sA\right)$ denote the continuous/discrete state-action space; $P_{s, s'}^a = P(s_{t+1} = s'| s_t = s, a_t = a)$ is the transition probability of next state $s'\in\sS$ given current state $s\in\sS$ and action $a\in\sA$, determined by the environment; $R_s^a = \mathbb{E}[r_t|s_t=s, a_t=a]$ returns the expected reward from the task when taking action $a$ on state $s$; $\mu_0(s)$ denotes the initial state distribution and $\gamma \in [0, 1)$ is a discounting factor. The effectiveness of a policy $\pi(a|s)$ is evaluated by its expected infinite-horizon reward: $\eta(\pi) = \mathbb{E}_{s_0 \sim \mu_0,a_t \sim \pi(s_t),s_{t+1} \sim P_{s_t,s_{t+1}}^{a_t}} \left[ \sum_{t=0}^\infty \gamma^t r_t \right]$.

\textbf{The Option Framework:}
Options $\sO=\{1,\cdots,K\}$ are introduced for modeling the policy-switching procedure on long-term tasks~\citep{SMDP}.
An option model is defined as a tuple $\gO=\left(\sS, \sA, \{\sI_o, \pi_o, \beta_o\}_{o\in \sO}, \pi_{\gO}(o|s), P_{s, s'}^a, R_s^a, \mu_0, \gamma\right)$, where, $\sS, \sA, P_{s, s'}^a, R_s^a, \mu_0, \gamma$ are defined as the same as MDP; $\sI_o \subseteq \sS$ denotes an initial set of states, from which an option can be activated; $\beta_o(s)= P(b=1|s)$ is a terminate function which decides whether current option should be terminated or not on a given state $s$; $\pi_o(a|s)$ is the intra-option policy that determines an action on a given state within an option $o$; a new option is activated in the call-and-return style by an inter-option policy $\pi_\gO(o|s)$ once the last option terminates.

\textbf{Generative Adversarial Imitation Learning (GAIL):}
For simplicity, we denote $(x_0, \cdots, x_n)$ as $x_{0:n}$ for short throughout this paper. Given expert demonstrations on a specified task as $ \sD_{demo} = \left\{ \tau_E=(s_{0:T}, a_{0:T})\right\}$, imitation learning aims at finding a policy $\pi$ that can best reproduce the expert's behavior, without the access of the real reward.
\citet{GAIL,fgail} cast the original maximum-entropy inverse reinforcement learning problem into an occupancy measurement \citep{ocm} matching problem:
\begin{align}
    \label{eq:gail}
    \argmin_{\pi} D_f\left(\rho_{\pi}(s, a)\| \rho_{\pi_E}(s, a)\right),
\end{align}
where, $D_f$ computes $f$-Divergence, $\rho_{\pi}(s,a)$ and $\rho_{\pi_E}(s,a)$ are the occupancy measurements of agent and expert respectively. By introducing a generative adversarial structure~\citep{gan}, GAIL minimizes the discrepancy
via alternatively optimizing the policy and the estimated discrepancy. To be specific, a discriminator $D_{\theta}(s, a): \sS \times \sA \mapsto (0, 1)$ parameterized by $\theta$ in GAIL is maximized and then the policy is updated to minimize the overall discrepancy along each trajectory of exploration. Such optimization process can be cast into: $\max_{\theta}\min_\pi \mathbb{E}_{\pi}\left[\log D_{\theta}(s, a)\right]+ \mathbb{E}_{\pi_E}\left[\log \left(1 - D_{\theta}(s, a)\right)\right]$.

\section{Our Framework: \method}
\label{sec:metho}
In this section, we provide the details of the proposed \method. Our goal is towards imitation learning from demonstrations of long-horizon tasks consisting of small sub-tasks. We first introduce necessary assumptions to facilitate our analyses, and follow it up by characterizing the motivations and formulation of \method. We then provide the EM-like algorithm towards the solution and conclude this section with theoretical analyses on the convergence of our algorithm.  

\subsection{Assumptions and Definitions}\label{sec:metho:define}
    We have introduced the full option framework for modeling switching procedure on hierarchical tasks in \Secref{sec:prelim}. However, it is inconvenient to directly learn the policy of this framework due to the difficulty of determining the initial set $\sI_o$ and break condition $\beta_o$. Here, we introduce the following assumption given which the original option framework will be equivalently converted into the one-step option that is easier to be dealt with. 

    \begin{assumption}[Valid Option Switching]\label{as:available_option}
        We assume that, 1) each state is switchable: $I_o=\sS, \forall o\in\sO$;  2) each switching is effective: $P(o_t=o_{t-1}|\beta_{o_{t-1}}(s_{t-1})=1)=0$.
    \end{assumption}
    Assumption~\ref{as:available_option} asserts that each state is switchable for each option, and once the switching happens, it switches to a different option with probability 1. Such assumption usually holds in practice without the sacrifice of model expressiveness. Now, we define the following one-step model.
    \begin{definition}[One-step Option]\label{de:one-step-opt}
        We define $\gO_{\text{one-step}}=\left(\sS, \sA, \sO^+, \pi_H, \pi_L, P_{s, s'}^a,  R_s^a, \tilde{\mu}_0, \gamma \right)$ where $\sO^+ = \sO\cup\{\#\}$ consists of all options plus a special initial option class satisfying $o_{-1} \equiv \#, \beta_\#(s) \equiv 1$. Besides, $\tilde{\mu}_0(s, o) \doteq \mu_0 (s) \mathds{1}_{o=\#}$, where $\mathds{1}_{x=y}$ is the indicator function, and it is equal to 1 iff $x=y$, otherwise 0. The high- and low-level policies are defined as:
        \begin{align}
        \label{eq:one-step-policy}
        \begin{split}
			\pi_H (o|s, o') &\doteq \beta_{o'}(s) \pi_{\gO}(o|s) + \left(1-\beta_{o'}(s)\right)\mathds{1}_{o = o'},\\
			\pi_L (a|s, o) &\doteq \pi_o(a|s).
		\end{split}\end{align}
    \end{definition}
    It can be derived that $\gO_{\text{one-step}}=\gO$ under Assumption~\ref{as:available_option}\footnote{The proofs of all theorems are provided in Appendix.}. This equivalence is beneficial as we can characterize the switching behavior by only looking at $\pi_H$ and $\pi_L$ without the need to justify the exact beginning/breaking condition of each option. We denote $\tilde{\pi} \doteq (\pi_H, \pi_L)$ and $\tilde{\Pi}=\{\tilde{\pi}\}$ as the set of stationary policies. We will employ the one-step option in the remainder of our paper. Note that in a current paper~\cite{Liskill} the rigorous notions and formulations have been provided for further discussing the relationship between the full option model and the one-step option model. 
    
    We also provide the definition of the option-occupancy measurement by borrowing the notion of the occupancy measurement $\rho(s,a)$ from MDP~\citep{ocm}.
    \begin{definition}[Option-Occupancy Measurement] \label{de:rho_s_a_o_o'}
        We define the option-occupancy measurement as $\rho_{\tilde{\pi}}(s, a, o, o') \doteq \mathbb{E}_{\tilde{\pi}} \left[\sum_{t=0}^\infty \gamma^t \mathds{1}_{(s_t=s,a_t=a,o_t=o,o_{t-1}=o')}\right]$.
    \end{definition}
    The measurement $\rho_{\tilde{\pi}}(s, a, o, o')$ can be explained as the distribution of the state-action-option tuple that generated by the policy $\pi_H$ and $\pi_L$ on a given $\tilde{\mu}_0$ and $P_{s, s'}^a$. According to the Bellman Flow constraint, one can easily obtain that $\rho_{\tilde{\pi}}(s,a,o,o')$ belongs to a feasible set of affine constraint $\sD=\{\rho(s, a, o, o')\geq 0; \sum_{a, o}\rho(s, a, o, o') = \tilde{\mu}_0 (s, o')+\gamma \sum_{s', a', o''} P_{s', s}^{a'} \rho(s', a', o', o'')\}$.

\subsection{Problem Formulation} \label{sec:metho:formu}
    Now we focus on the imitation task on long-term demonstrations. Note that GAIL is no longer suitable for this scenario since it is hard to capture the hierarchy of sub-tasks by MDP. Upon GAIL, the natural idea is to model the long-term tasks via the one-step Option instead of MDP, and the policy is learned by minimizing the discrepancy of the occupancy measurement between expert and agent, namely,
    \begin{eqnarray}
    \label{eq:gail-goal}
    \min_{\tilde{\pi}} D_f\left(\rho_{\tilde{\pi}}(s, a)\| \rho_{\tilde{\pi}_E}(s, a)\right).
    \end{eqnarray}
    While this idea is straightforward and never explored before, we claim that it will cause solution ambiguity---we cannot make sure that $\tilde{\pi}=\tilde{\pi}_E$, even we can get the optimal solution $\rho_{\tilde{\pi}}(s, a)=\rho_{\tilde{\pi}_E}(s, a)$ in \Eqref{eq:gail-goal}. 
    
    Intuitively, for the hierarchical tasks, the action we make depends not only on the current state we observe but also on the option we have selected. By Definition~\ref{de:one-step-opt}, the hierarchical policy is relevant to the information of current state, current action, last-time option and current option, which motivates us to leverage the option-occupancy measurement instead of conventional occupancy measurement to depict the discrepancy between expert and agent. Actually, we have a one-to-one correspondence between $\tilde{\Pi}$ and $\sD$.
    \begin{theorem}[Bijection] \label{th:bijection}
    For each $\rho\in\sD$, it is the option-occupancy measurement of the following policy:
    \begin{align}
        \pi_H=\frac{\sum_{a}\rho(s, a, o, o')}{\sum_{a, o}\rho(s, a, o, o')}, \pi_L=\frac{\sum_{o'}\rho(s, a, o, o')}{\sum_{a, o'}\rho(s, a, o, o')},
    \end{align}
    and $\tilde{\pi}=(\pi_H,\pi_L)$ is the only policy whose option-occupancy measure is $\rho$.
    \end{theorem}

    With Theorem~\ref{th:bijection}, optimizing the option policy is equivalent to optimizing its induced option-occupancy measurement, since $\rho_{\tilde{\pi}}(s, a, o, o') = \rho_{\tilde{\pi}_E}(s, a, o, o') \Leftrightarrow \tilde{\pi} = \tilde{\pi}_E$. Our hierarchical imitation learning problem becomes:
    \begin{align} \label{eq:overall-goal}
        \min_{\tilde{\pi}} D_f\left(\rho_{\tilde{\pi}}(s, a, o, o')\| \rho_{\tilde{\pi}_E}(s, a, o, o')\right)
    \end{align}
    
\begin{figure}[t!] \begin{center}
        \includegraphics[width=0.95\linewidth]{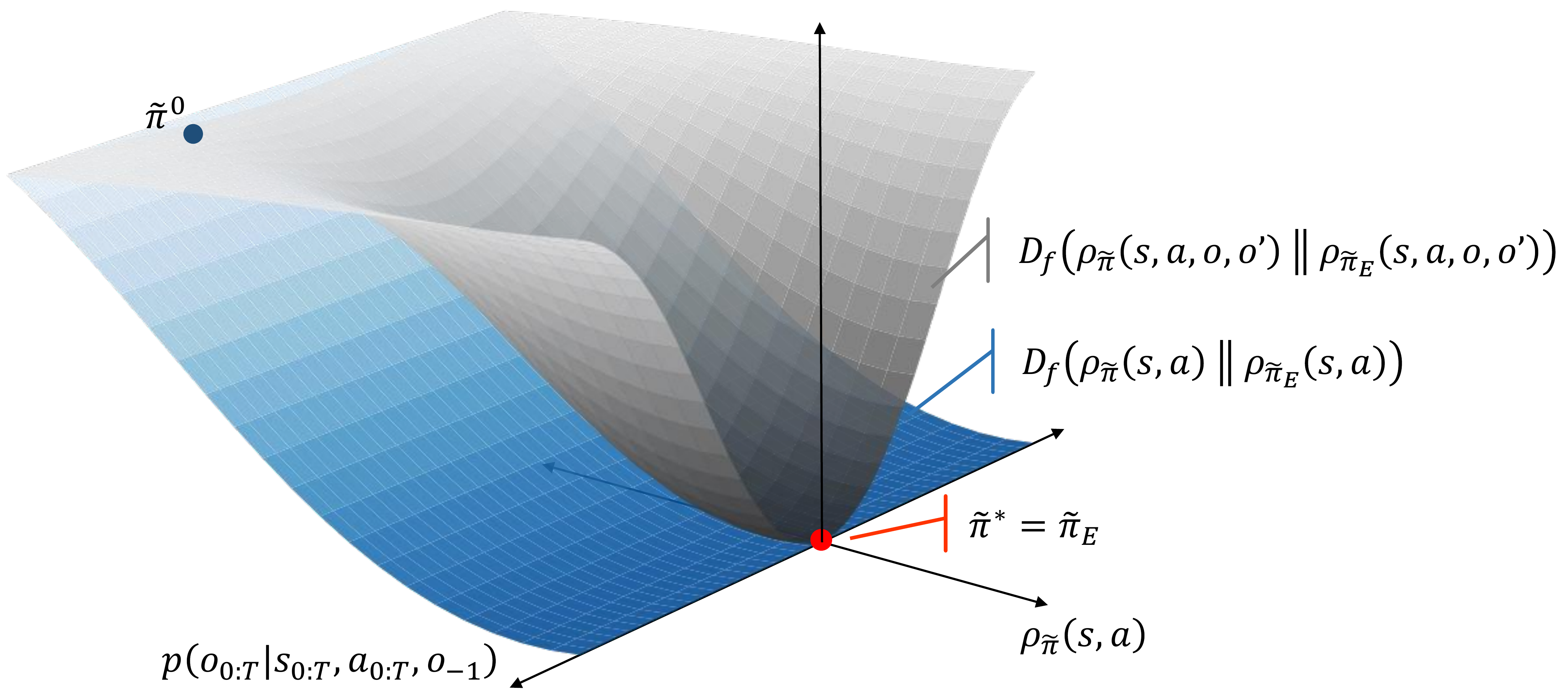} 
        \caption{Illustration of the different convergence properties of the optimisation problems defined in \Eqref{eq:gail-goal} and \Eqref{eq:overall-goal}. The options are not explicitly constrained by \Eqref{eq:gail-goal}}.
        \label{fig:convergence}
    \end{center} \end{figure}

    Note that the optimization problem defined in \Eqref{eq:overall-goal} implies that in \Eqref{eq:gail-goal}, but not vice versa: (1) since $\rho_{\tilde{\pi}} (s, a) = \sum_{o, o'} \rho_{\tilde{\pi}} (s, a, o, o')$, we can easily obtain $\rho_{\tilde{\pi}} (s, a, o, o')= \rho_{\tilde{\pi}_E} (s, a, o, o')\Rightarrow \rho_{\tilde{\pi}}(s, a)= \rho_{\tilde{\pi}_E}(s, a)$; (2) as $ \mathbb{E}_{q(o,o')} \left[ D_f\left(\rho_{\tilde{\pi}}(s, a, o, o')\|\rho_{\tilde{\pi}_E}(s, a, o, o')\right)\right] \geq \mathbb{E}_{q(o,o')} \left[D_f\left(\rho_{\tilde{\pi}}(s, a)\|\rho_{\tilde{\pi}_E}(s, a)\right)\right]$ for any option distribution $q(o,o')$, addressing the problem defined in \Eqref{eq:overall-goal} is addressing an upper bound of that defined in \Eqref{eq:gail-goal}. Indeed, we will show in \textsection~\ref{sec:metho:analysis} that decreasing the goal in \Eqref{eq:overall-goal} will definitely decrease that of \Eqref{eq:gail-goal} given certain conditions. Figure~\ref{fig:convergence} depicts the relationship between \Eqref{eq:overall-goal} and \Eqref{eq:gail-goal}.
    
    Yet, it is nontrivial to tackle \Eqref{eq:overall-goal} particularly owing to the unobserved options in expert demonstrations. Here, we propose an EM-like algorithm to address it, which will be detailed in \textsection~\ref{sec:metho:ocm-matching} and \textsection~\ref{sec:metho:opt-viterbi}. 
 
\subsection{Option-Occupancy Measurement Matching}\label{sec:metho:ocm-matching}
    In this section, we assume the expert options are given and
    train the hierarchical policy $\tilde{\pi}$ to minimize the goal defined in \Eqref{eq:overall-goal}. We denote the option-extended expert demonstrations as $\tilde{\sD}_\text{demo} = \{\tilde{\tau}=(s_{0:T}, a_{0:T}, o_{-1:T})\}$. 
    
    Inspired from GAIL~\citep{GAIL}, rather than calculating the exact value of the option-occupancy measurement, we propose to estimate the discrepancy by virtue of adversarial learning. We define a parametric discriminator as $D_{\theta}(s, a, o, o'): \sS \times \sA \times \sO \times \sO^+ \mapsto (0, 1)$. If specifying the f-divergence as Jensen–Shannon divergence, \Eqref{eq:overall-goal} can be converted to a min-max game:
    \begin{align} \begin{split}
    \label{eq:opt-gail-goal}
     \min_{\tilde{\pi}} \max_{\theta} \:&\mathbb{E}_{\tilde{\sD}_\text{demo}}\left[\log \left(1-D_{\theta}(s, a, o, o')\right)\right] \\
        & \quad + \mathbb{E}_{\tilde{\sD}_{\tilde{\pi}}}\left[\log \left(D_{\theta}(s, a, o, o')\right)\right].
    \end{split} \end{align}
    The inner loop is to train $D_{\theta}(s, a, o, o')$ with the expert demonstration $\tilde{\sD}_\text{demo}$ and samples $\tilde{\sD}_{\tilde{\pi}}$ generated by self-exploration. Regarding the outer loop, a hierarchical reinforcement learning (HRL) method should be used to minimize the discrepancy:
    \begin{align}
        &\tilde{\pi}^\star = \argmin_{\tilde{\pi}} \mathbb{E}_{\tilde{\pi}}\left[c(s, a, o, o')\right] - \lambda \mathbb{H}(\tilde{\pi})\label{eq:hrl-goal},
    \end{align}
    where $c(s, a, o, o') = \log D_{\theta}(s, a, o, o')$ and the causal entropy $\mathbb{H}(\tilde{\pi}) = \mathbb{E}_{\tilde{\pi}}\left[-\log \pi_H - \log \pi_L\right]$ as a policy regularizer with $\lambda \in [0, 1]$. 
    Notice that the cost function is related to options, which is slightly different from many HRL problems with option agnostic cost/reward~\citep{Option-AC, DAC}. 
    To deal with this case, we reformulate the option-reward and optimize \Eqref{eq:hrl-goal} using similar idea as \citet{DAC}. 
    
    DAC characterizes the option model as two-level MDPs. Here we borrow its theoretical result by re-interpreting the reward used by each MDP: For the high-level MDP, we define state $s_t^H \doteq (s_t, o_{t-1})$, action $a_t^H \doteq o_t$, and reward $R^H((s, o'), o) \doteq -\sum_{a}\pi_L(a_t|s_t, o_t)c(s_t, a_t, o_t, o_{t-1})$. 
    For the low-level MDP, we denote $s_t^L \doteq (s_t, o_t)$, $a_t^L \doteq a_t$, and $R^L((s, o), a) = -\sum_{o'} c(s, a, o, o')p_{\pi_H}(o'|s, o)$ with the posterior propability $p_{\pi_H}(o'|s, o) = \pi_H(o|s, o')\frac{p(o'|s)}{p(o|s)}$. Other terms including the initial state distributions $\mu_0^H$ and $\mu_0^L$, the state-transition dynamics $P^H$ and $P^L$ are defined similar to \citet{DAC}. The HRL task in \Eqref{eq:hrl-goal} can be separated into two non-hierarchical ones with augmented MDPs: $\sM^H = (S^H, A^H, P^H, R^H, \mu_0^H, \gamma)$, $\sM^L = (S^L, A^L, P^L, R^L, \mu_0^L, \gamma)$, whose action decisions depend on $\pi_H$ and $\pi_L$, separately. Such two non-hierarchical problems can be solved alternatively by utilizing typical reinforcement learning methods like PPO~\citep{PPO} with immediate imitation reward $r_t = -c(s_t, a_t, o_t, o_{t-1})$ when in practice.
    
    By alternating the inner loop and the outer loop, we finally derive $\tilde{\pi}^\star$ that addresses \Eqref{eq:overall-goal}.

\subsection{Expert Option Inference}\label{sec:metho:opt-viterbi}
     So far, we have supposed that the expert options are provided, which, however, is usually not true in practice. It thus demands a kind of method to infer the options from observed data (states and actions). Basically, given a policy, the options are supposed to be the ones that maximize the likelihood of the observed state-actions, according to the principle of Maximum-Likelihood-Estimation (MLE). 
    
     We approximate the expert policy with $\tilde{\pi}$ learned by the method above. With states and actions observed, the option model will degenerate to a Hidden-Markov-Model (HMM), therefore, the Viterbi method~\citep{viterbi} is applicable for expert option inference. 
     We call this method as Option-Viterbi for its specification to the option model. 
    
    Given current learned $\pi_H, \pi_L$ and $\tau = (s_{0:T}, a_{0:T})\in \sD_\text{demo}$, Option-Viterbi generates the most probable values of $o_{-1:T}$ that induces the maximization of the whole trajectory:
    \begin{align} \begin{split}\label{eq:opt-viterbi}
	    	& \argmax_{o_{-1:T}} P(s_{0:T}, a_{0:T}, o_{-1:T})\\
          \propto& \argmax_{o_{0:T}} \log P(a_{0:T}, o_{0:T} | s_{0:T}, o_{-1}=\#)\\
          = &\argmax_{o_T} \hat{\alpha}_T(o_T).
	\end{split} \end{align}
	Here a maximum foreword message $\hat{\alpha}_t(o_t)$ is introduced for indicating the maximum probability value of the partial trajectory till time $t$ given $o_t$, and it can be calculated recursively as \Eqref{eq:opt-viterbi-forward}:
	
	\begin{gather} \label{eq:opt-viterbi-forward} \begin{aligned} 
	\begin{split}
	\hat{\alpha}_t(o_t) =& \max_{o_{o:t-1}} \log P(a_{0:t}, o_t| s_{0:t}, o_{0:t-1}, o_{-1}=\#)\\
	=& \max_{o_{t-1}} \hat{\alpha}_{t-1}(o_{t-1}) + \log \pi_H(o_t | s_t, o_{t-1})\\
		   &\qquad + \log \pi_L(a_t| s_t, o_t),
	\end{split}
	\end{aligned}\\ 
	\hat{\alpha}_0(o_0) = \log \pi_L(a_0| s_0, o_0) + \log \pi_H(o_0 | s_0, \#).
	\end{gather}

	Clearly, Option-Viterbi has a computation complexity of $O(T\times K^2)$ for a $T$-step trajectory with $K$ options. By back-tracing $o_{t-1}$ that induces the maximization on $\hat{\alpha}(o_t)$ at each time step, the option-extended expert trajectories $ \tilde{\sD}_\text{demo}$ can finally be acquired. 

    Although our above option inference process implies that the expert demonstrations are assumed to be generated through a hierarchical policy as the same as the agent, our method is still applicable to the non-hierarchical expert, given the fact that a non-hierarchical expert can be imitated by a hierarchical agent (with fewer options activated).
  
\subsection{Summary of \method} \label{sec:metho:overview}
	We briefly give an overview of our proposed \method in Algorithm~\ref{alg}.
	With expert-provided demonstrations $\sD_\text{demo}=\{\tau_E=(s_{0:T}, a_{0:T})\}$ and initial policy $\tilde{\pi}$, our method alternates the policy optimization and option inference for sufficient iterations. 
    \begin{algorithm}
    	\caption{\method}\label{alg}
    	\SetAlgoLined
    	\KwData{Expert trajectories $\sD_\text{demo}=\left\{\tau_E=\left\{(s_{0:T}, a_{0:T})\right\}\right\}$}
    	\KwIn{Initial policy $\tilde{\pi}^0=(\pi_H^0,\pi_L^0)$}
    	\KwOut{Learned Policy $\tilde{\pi}^\star$}
    	\For{$n = 0 \cdots N$} {~
    		\textbf{E-step:} Infer expert options with $\tilde{\pi}^n$ by (\ref{eq:opt-viterbi-forward}) and get $\tilde{\sD}_\text{demo}$; sample trajectories with $\tilde{\pi}^n$ and get $\tilde{\sD}_{\tilde{\pi}}$.\\
    		\textbf{M-step:} Update $\tilde{\pi}^{n+1}$ by (\ref{eq:opt-gail-goal}).
    	}
    \end{algorithm}

\subsection{Convergence Analysis} \label{sec:metho:analysis}
    Algorithm~\ref{alg} has depicted how our method is computed, but it is still unknown if it will converge in the end. To enable the analysis, we first define a surrogate distribution of the options as $q(o_{-1:T})$. 
    We denote $Q_n$ as the expected objective in \Eqref{eq:overall-goal} at iteration $n$ by Algorithm~\ref{alg}, namely, $Q_n=\mathbb{E}_{q(o_{-1:T})}\left[D_f\left(\rho_{\tilde{\pi}}(s, a, o, o')\| \rho_{\tilde{\pi}_E}(s, a, o, o')\right)\right]$. We immediately have the following convergence theorem for Algorithm~\ref{alg}.
    
    \begin{theorem}[Convergence]
    \label{th:convergence}
    Algorithm~\ref{alg} only decreases the objective: $Q_{n+1}\leq Q_n$, if these two conditions hold: (1) $q(o_{-1:T})$ is our option sampling strategy in E-step and is equal to the posterior of the options given current policy $\tilde{\pi}^n$, \emph{i.e.} $q(o_{-1:T})=P_{\tilde{\pi}^n}(o_{-1:T}|s_{0:T},a_{0:T})$; (2) M-step definitely decreases the objective in \Eqref{eq:overall-goal} at each iteration.
    \end{theorem}
   
    The proof is devised by generalizing traditional EM algorithm to the f-divergence minimization problem. We provide the details in Appendix. Since $Q_n\geq 0$, Theorem~\ref{th:convergence} confirms that Algorithm~\ref{alg} will converge eventually. The first condition in Theorem~\ref{th:convergence} guarantees that the objective of \Eqref{eq:gail-goal} is equal to $Q_n$ after E-step, thus Algorithm~\ref{alg} also helps minimize \Eqref{eq:gail-goal}. Besides, if the global minimum is achieved, we have $Q_n=0\Rightarrow \rho_{\tilde{\pi}}(s, a, o, o')=\rho_{\tilde{\pi}_E}(s, a, o, o')\Rightarrow \tilde{\pi}=\tilde{\pi_E}$. 
   
    In our implementation, as mentioned in \textsection~\ref{sec:metho:opt-viterbi}, we adopt the maximized sampling process instead of the posterior sampling that is required by Theorem~\ref{th:convergence}, as, in this way, we find that it still maintains the convergence in our experiments while reducing the computation complexity with only sampling one option series per trajectory.

\section{Experiments}
\label{sec:exp}
\begin{table*}[htbp]
    \label{tab: comp}  \begin{center}
    \caption{Comparative results. All results are measured by the average \textbf{maximum average reward-sum} among different trails.}
    \begin{tabular}{ccccc}
    \toprule 
    & \small{Hopper-v2}& \small{Walker2d-v2} & \small{AntPush-v0} &\small{CloseMicrowave2} \\
    \midrule
    Demos $(s, a)\times T$ & $(\mathbb{R}^{11}, \mathbb{R}^3) \times 1$k & $(\mathbb{R}^{17}, \mathbb{R}^6) \times 5$k & $(\mathbb{R}^{107}, \mathbb{R}^8) \times 50$k & $(\mathbb{R}^{101}, \mathbb{R}^8) \times 1$k\\
    \small{Demo Reward} & 3656.17$\pm$0.0 & 5005.80$\pm$36.18 & 116.60$\pm$14.07 & 149.68 $\pm$ 16.29 \\
    \midrule
    \small{BC} & 275.93$\pm$31.09 & 589.17$\pm$90.81 & 4.60$\pm$2.72 & 26.03$\pm$2.33\\
    \small{GAIL} & 535.29$\pm$7.19 & 2787.87$\pm$2234.46 & 54.82$\pm$4.81 & 39.14$\pm$12.88\\
    \midrule 
    \small{H-BC} & 970.91$\pm$72.69 & 3093.56$\pm$107.11 & 89.23$\pm$1.37 & 89.54$\pm$15.44\\
    \small{GAIL-HRL} & 3697.40$\pm$1.14 & 3687.63$\pm$982.99 & 20.53$\pm$6.89 & 56.95$\pm$25.74\\
    \small{Ours} & \textbf{3700.42$\pm$1.70} & \textbf{4836.85$\pm$100.09} & \textbf{95.00$\pm$2.70} & \textbf{100.74$\pm$21.33}\\
    \bottomrule 
    \end{tabular}
\end{center} \end{table*}
\begin{figure*}[t!] \begin{center}
    \includegraphics[width=0.95\linewidth]{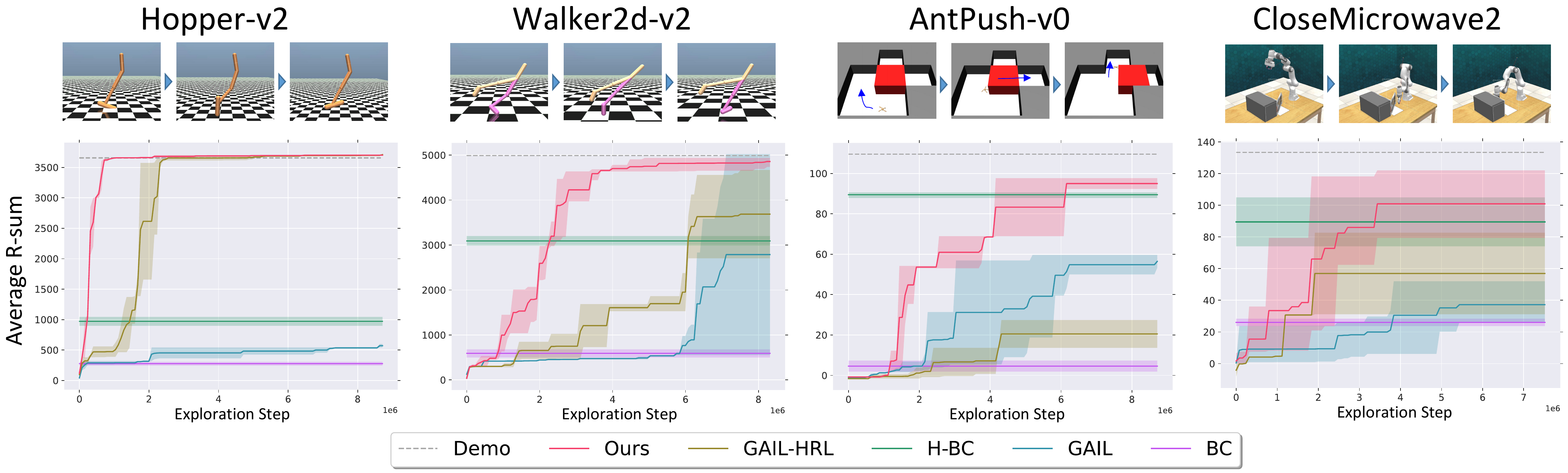}
    \caption{comparison of learning performance on four environments. The environments and the task designs are illustrated on the top of the figure with more details provided in \textsection~\ref{sec:exp:env}. 
    We compare the \textbf{maximum average reward-sums} vs. exploration steps on different environments. The solid line indicates the average performance among several trials under different random seeds, while the shade indicates the range of the maximum average reward-sums over different trials.}
    \label{fig:result_compare}
\end{center} \end{figure*}

We evaluate our model on four widely used robot learning benchmarks with locomotion and manipulation tasks. We first compare our model against other counterparts that do not employ hierarchical structure or self-explorations.  Then we provide an ablated study to understand the impact of each component in our model.

\subsection{Environments} \label{sec:exp:env}
Four environments are used for evaluations:

\textbf{Hopper-v2} and \textbf{Walker2d-v2:} The Hopper-v2 and the Walker2d-v2 are two standardized continuous-time locomotion environments implemented in the OpenAI Gym~\citep{gym} with the MuJoCo~\citep{mujoco} physics simulator. On these two tasks, the robot should move toward the desired direction by moving its leg periodically. We use expert demonstration containing 1,000 time-steps for learning on Hopper-v2 environment and 5,000 time-steps for learning on the Walker2d-v2 environment. Both of the expert demonstrations are generated by a policy trained by PPO~\citep{PPO}.

\textbf{AntPush-v0:} The AntPush-v0 is a navigation/locomotion task proposed in \citet{hiro}, where an Ant robot is required to navigate into a room that is blocked by a movable obstacle, as shown in Figure~\ref{fig:result_compare}. Specifically, the robot should first navigate to and push away the obstacle and then go into the blocked room. For better comparing the learning performance, we slightly extend the original binary reward as $r_t = -\Delta \|\text{pos}_t-\text{tar}\|_2^2 + 100 \times \mathds{1}_{\text{pos}_t = \text{tar}}$ where $\text{pos}$ is the position of the robot and $\text{tar}$ is the location of the blocked room.
We use expert demonstrations containing 50,000 time-steps for learning in this environment, where the policy is trained with DAC~\citep{DAC} regarding our modified reward.

\textbf{CloseMicrowave2:} The Closemicrowave2 is a more challenging robot operation environment in RLBench~\citep{rlbench}. A 6-DoF robot arm with a gripper is required to reach and close the door of a microwave by controlling the increments on joint positions of the robot arm, as shown in Figure~\ref{fig:result_compare}.
The reward is defined as $r_t = -\Delta \theta_t + 100 \times \mathds{1}_{\theta_t = 0}$, where $\theta$ denotes the rotation angle of the door,
We use expert demonstrations containing 1,000 time-steps for learning in this environment generated by a handcrafted feedback controller.

\subsection{Comparative Evaluations} \label{sec:exp:comp}

To illustrate the effectiveness of the proposed method, we contrast several popular imitation learning baselines, including:
    1) \textbf{supervised Behavior Cloning (BC)}~\citep{bc1} that do not contain hierarchical structure or self-exploration; 
    2) \textbf{GAIL}~\citep{GAIL} that uses self-exploration without hierarchical structure; 
    3) \textbf{Hierarchical Behavioral Cloning (H-BC)}~\citep{provable_h-bc}, which builds upon the Option structure, but optimizing both the high- and low-level policies by directly maximizing the probability of the observed expert trajectories without any self-exploration. This baseline can also be regarded as optimizing \Eqref{eq:opt-viterbi-forward} with forward-backward messages; 
    4) a variant of GAIL, denoted as \textbf{GAIL-HRL} that updates hierarchical policies according to \Eqref{eq:gail-goal}, where the immediate imitation reward $r_t = -\log D_{\theta'}(s_t, a_t)$ is used instead. GAIL-HRL can be regarded as a simplified version of our \method without using options in occupancy measurement matching, and it is designed for justifying the necessity of involving options during the whole occupancy measurement matching procedure. 
    
    We employ the same demonstration trajectories, network structures, and option numbers on each environment for fair comparisons. Specifically, we allow 4 available option classes for all environments, a Multi-Layer Perception(MLP) with hidden size (64, 64) to implement the policies of both levels on Hopper-v2, Walker2d-v2, AntPush-v0, and (128, 128) on Closemicrowave2; the discriminator is realized by an MLP with hidden size (256, 256) on all environments. For methods that do not use self-exploration, we train the policy for 100 epochs and then evaluate it by average reward-sums; for methods that rely on self-exploration, we update the policy and record the average reward-sum every 4096 environmental steps, and record maximum average reward-sums over the whole training period. The evaluation results are displayed in Figure~\ref{fig:result_compare}.
    
    \begin{figure} \begin{center}
        \includegraphics[width=0.9\linewidth]{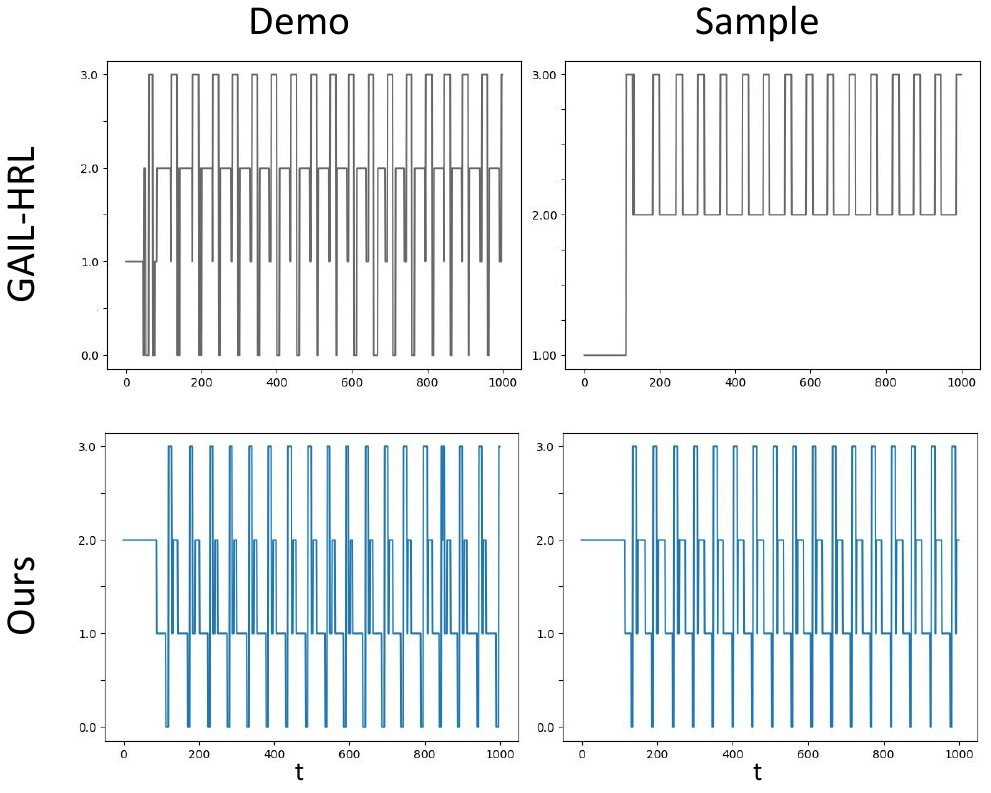}
        \caption{Visualization of the options activated at each step, learned respectively by our proposed method (blue) and GAIL-HRL (gray) on Walker2d-v2. 'Demo' denotes the options inferred from the expert, and 'Sample' denotes the options used by agent when doing self-explorations. The effectiveness of our proposed method on regularizing the option switching is obvious by comparing the consistent switching tendencies between Demo and Sample.}
        \label{fig:result_detail_opt}
    \end{center} 
    \end{figure}

Obviously, our method gains faster convergence speed and better eventual performance than all baselines in general, as illustrated in Figure~\ref{fig:result_compare}. For example, on CloseMicrowave2, which is clearly a long-horizon task, our \method converges to a remarkably larger reward than all others. Besides, by introducing the option-based hierarchical structure, \method, H-BC and GAIL-HRL are superior to the counterparts that use single-level policy, namely, BC and GAIL. When demonstrations are limited, for instance, on Hopper-v2, Walker2d-v2, and Closemicrowave2, GAIL-like methods including \method and GAIL-HRL outperform their BC-like counterpart H-BC, probably thanks to the reduction of compounding errors by self-explorations. On AntPush-v0, the advantage of our method over H-BC is reduced. We conjecture this is because H-BC is data-hungry, and it can become better when sufficient demonstrations are fed. Comparing with GAIL, with the help of hierarchical modeling, our method can successfully imitate the behavior of the expert with fewer self-explorations.

To examine if our proposed method actually regularizes the options switching between expert and agent, Figure~\ref{fig:result_detail_opt} illustrates the options including that are inferred from expert and that generated by agent on Walker2d-v2 with more examples deferred to Appendix. It is observed in our method that the expert's options are consistent with the agent's, while for GAIL-HRL, a dramatic conflict exists between expert and agent. This result confirms the benefit of our method since we explicitly minimize the discrepancy of the option-extended occupancy measurement between expert and agent in \Eqref{eq:overall-goal}.

\subsection{Ablations and Analysis} \label{sec:exp:ablation}
In this section,  we perform an in-depth analysis on each component of our model. 
    
    \begin{figure} \begin{center}
        \includegraphics[width=0.99\linewidth]{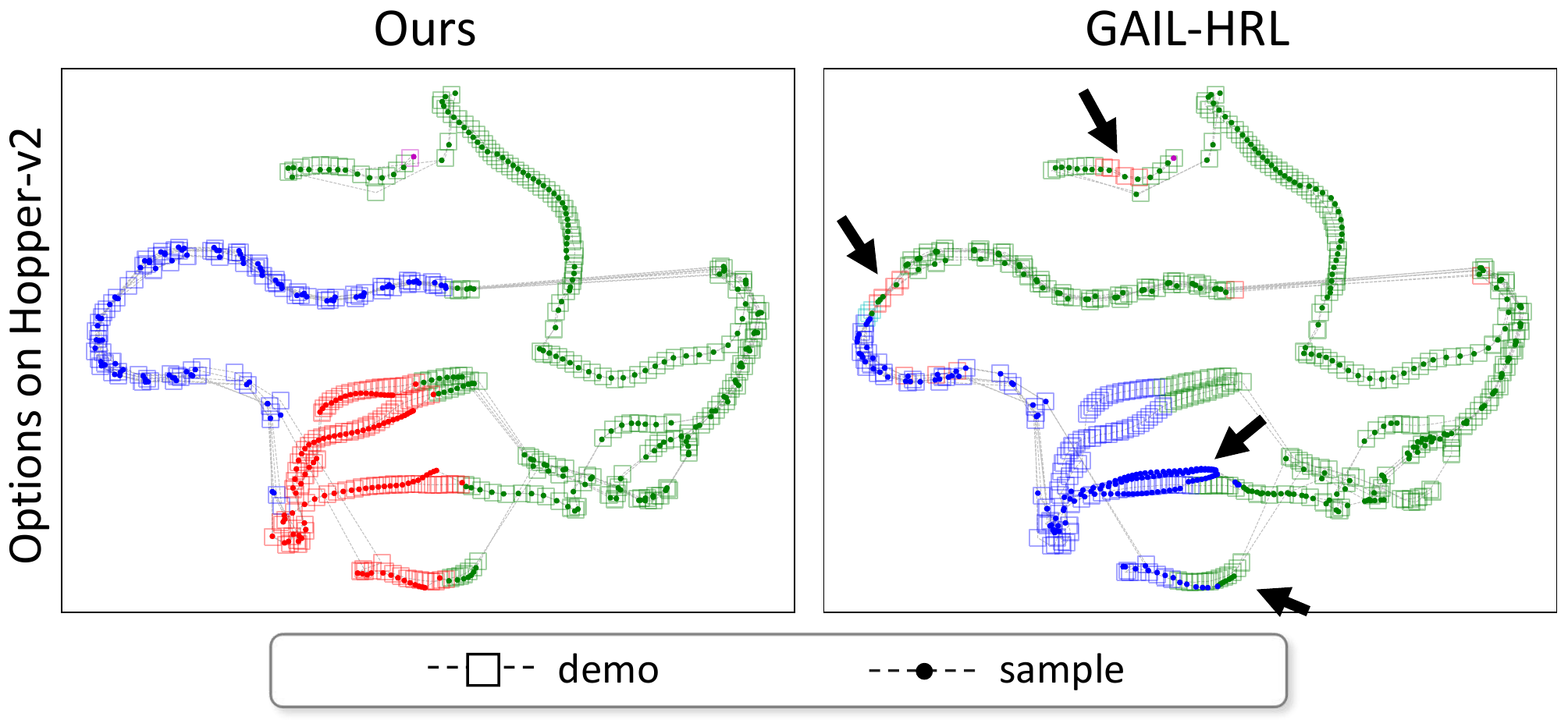}
        \caption{Visualization of the state-action and options on Hopper-v2 environment by t-SNE~\citep{tsne}. The dots and squares in the figure indicate the state-action pairs generated by agent and expert on each time step, respectively, and the gray lines connect time-adjacent points. The color indicates the option activated at each time-step. Arrows point to the miss-matched option switching behaviors between agent and expert demonstration by GAIL-HRL.}
        \label{fig:result_cluster}
    \end{center} \end{figure}

\textbf{The necessity of using option-extended occupancy:}
We geometrically assess the difference between our \method and GAIL-HRL on Walker2d-v2. To do so, we visualize the trajectories of the expert (squares) and the agent (dots) as well as their options (color) at each time step. Different colors indicate the different options used by the low-policy, and the expert's options are inferred with the agent's policy. The visualizations in Figure~\ref{fig:result_cluster} read that the agent's trajectory by \method almost overlaps with the demonstration, whereas the one by GAIL-HRL sometimes tracks a wrong direction with respect to the expert. Moreover, in \method, the options of the expert and agent always switch collaboratively with each other, while GAIL-HRL occasionally derives inconsistent options (highlighted by an arrow) between the expert and agent. This phenomenon once again verifies the necessity of performing the option-extended occupancy measurement matching in \Eqref{eq:overall-goal}. If otherwise conducting the occupancy measurement matching in \Eqref{eq:gail-goal}, it will give rise to ambiguities on the learned policies and eventually affect the learning performance.
    
\begin{figure} \begin{center}
    \includegraphics[width=0.99\linewidth]{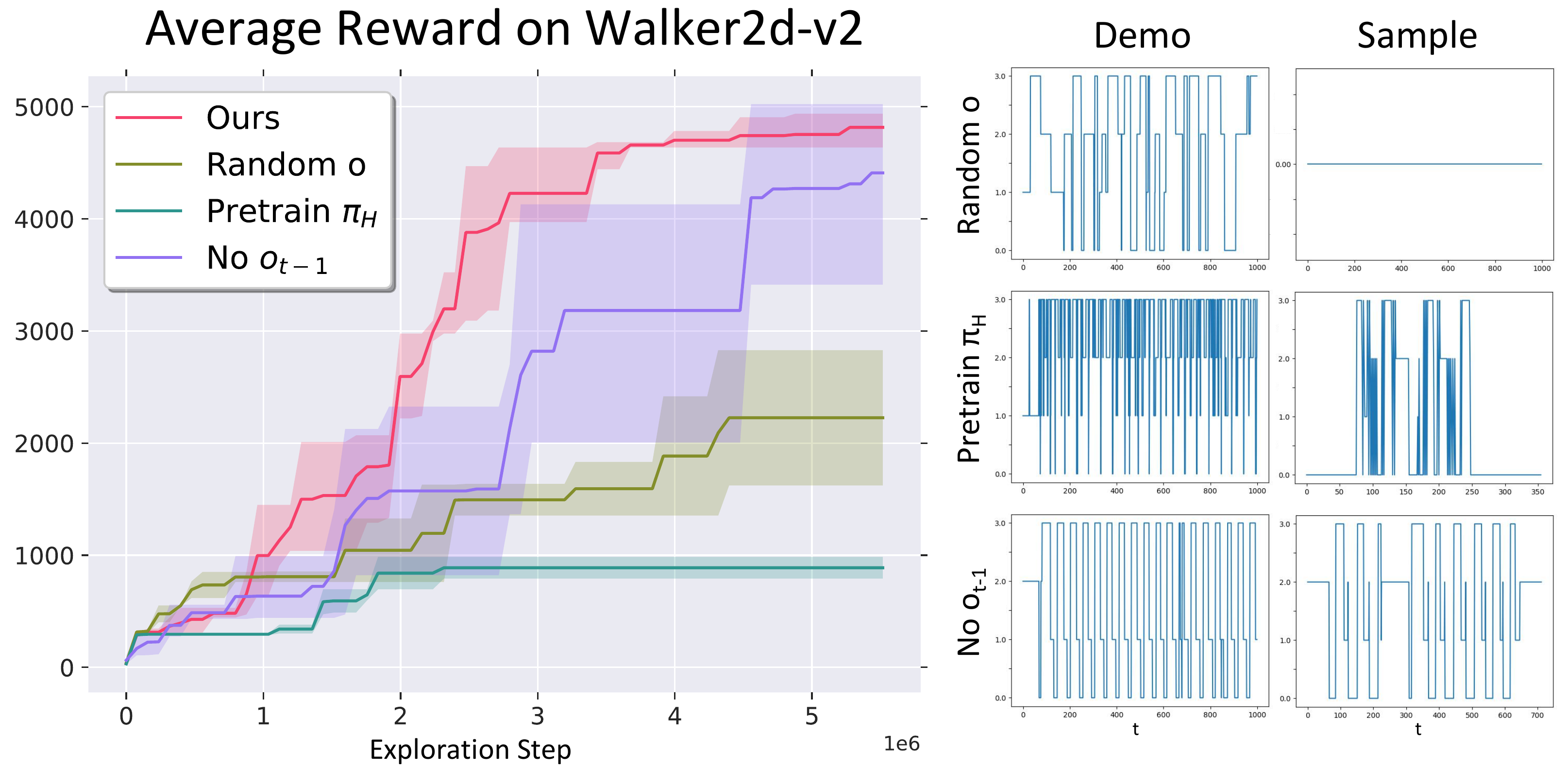}
    \caption{\label{fig:result_ablation}Ablations on individual components of our proposed method. The curve on the left side compares the \textbf{maximum average reward-sum} within a given range of exploration steps. The options used by agent and inferred from expert are presented one the right side.}
\end{center} \end{figure}

\textbf{Ablations on individual components:} 
To verify the rationality of the model design, we test different variants of our method in Figure~\ref{fig:result_ablation}.

\textbf{1) Random $o$.} In this case, we generate the expert's options by random without option inference. As observed, the random options mislead policy learning and thus yield worse performance. We further display the option switching in the right sub-figure in Figure~\ref{fig:result_ablation}. The high-level policy tends to be conservative on switching since the sampling options always keep unchanged, making our model less expressive to fit the expert's hierarchical policy. 

\textbf{2) No $o_{t-1}$.} 
we simply omit the $o_{t-1}$ in \Eqref{eq:opt-gail-goal} and implement the discriminator with only $(s_t, a_t, o_t)$. Clearly, without the option information from the last step, the high-level policy is not fully regularized and cannot capture the option dynamics, thereby delivering a performance drop compared to the full-option version. 

\textbf{3) Pretrain $\pi_H$.}
We pre-train the high-level policy using H-BC for 50 epochs and then fix it during the subsequent learning procedure, which can be regarded as a version of the two-stage method by \citet{d-info-gail}. Such ablation explains it is indeed demanded to simultaneously learn the high- and low-level policies end-to-end by examining the results in Figure~\ref{fig:result_ablation}. In the optimization perspective, the two-level policies are coupled with each other and will be improved better under an alternative training fashion.

In summary, all the results here support the optimal choice of our design. 

\begin{figure}[h] \begin{center}
    \includegraphics[width=0.99\linewidth]{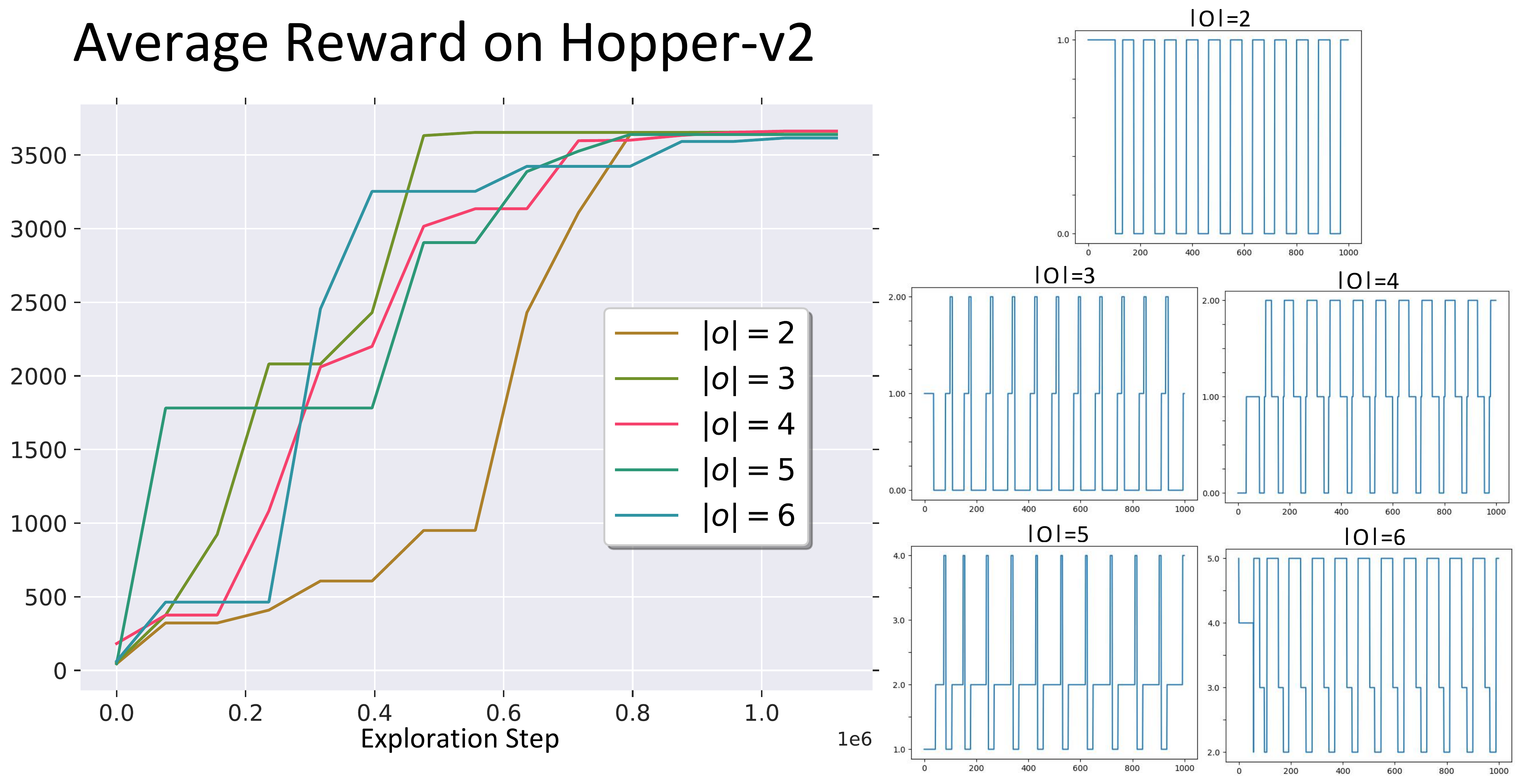}
    \caption{\label{fig:result_ablation_dif_o}Ablations on the total number of available option classes. Left: how different value of $|o|$ influences the \textbf{maximum average reward-sum} within a given range of exploration steps on Hopper-v2 environment. Right: actual option transition for different $|o|$.}
\end{center} \end{figure}
\textbf{Ablations on the number of available option classes:}
Throughout our above experiments, we set the number of option classes $|o|=4$ and find it works generally. In fact, changing $|o|$ does not essentially change our performance if $|o|$ is sufficiently large, as our algorithm will automatically inactivate redundant options (similar sub-task switching will be clustered into the same option class). 
For illustrating such robustness, we evaluate our method with $|o|\in\{2\cdots6\}$ on the Hopper-v2 environment. As observed from Figure\ref{fig:result_ablation_dif_o}, 
when $|o|\geq 3$, all variants share similar convergence behavior and option transition: the number of activated options is
observed as 3 for all cases. When $|o|\geq 2$, the convergence becomes worse, probably due to its less expressivity.

\section{Discussion and Conclusion}
\label{sec:conclusion}
In this paper, we have presented \method, a theoretically-grounded framework that integrates hierarchical modeling with option occupancy measurement matching. We then provide an EM-like algorithm for training the policies of \method with a provable guaranteed training convergence. Comparing to existing HIL methods, \method can regularize both the high- and low-level policies to better fit the hierarchical behavior of the expert. Experiments on robotic locomotion and manipulation benchmarks demonstrate the effectiveness of the proposed \method over other counterparts; \method works well particularly for the tasks consisting of evidently separable sub-tasks. Our method is generally powerful and can be enhanced by absorbing external knowledge. For example, it could be a future work direction to consider a long-horizon plan as the prior knowledge when inferring options.

\section*{Acknowledgement}
This research was primary funded by National Key R\&D Program of China (Grant No.20191241501). It was also partially funded by THU-Bosch JCML center (Grant No.20193000122). 
Meanwhile, this work is jointly sponsored by CAAI MindSpore Open Fund, China Postdoctoral Science Foundation (Grant No.2020M670337), and the National Natural Science Foundation of China (Grant No. 62006137).
We would like to thank Chao Yang and Yu Luo for their generous help and insightful advice.


\bibliography{IEEEabrv,references}  

\begin{thebibliography}{31}
\providecommand{\natexlab}[1]{#1}
\providecommand{\url}[1]{\texttt{#1}}
\expandafter\ifx\csname urlstyle\endcsname\relax
  \providecommand{\doi}[1]{doi: #1}\else
  \providecommand{\doi}{doi: \begingroup \urlstyle{rm}\Url}\fi

\bibitem[Atkeson \& Schaal(1997)Atkeson and Schaal]{bc2}
Atkeson, C.~G. and Schaal, S.
\newblock Robot learning from demonstration.
\newblock In \emph{Proc. Int. Conf. Mach. Learn.}, 1997.

\bibitem[Bacon et~al.(2017)Bacon, Harb, and Precup]{Option-AC}
Bacon, P.-L., Harb, J., and Precup, D.
\newblock The option-critic architecture.
\newblock In \emph{Proc. {AAAI} Conf. Artificial Intell.}, 2017.

\bibitem[Baum et~al.(1972)]{BW-alg}
Baum, L.~E. et~al.
\newblock An inequality and associated maximization technique in statistical
  estimation for probabilistic functions of markov processes.
\newblock \emph{Inequalities}, 3\penalty0 (1):\penalty0 1--8, 1972.

\bibitem[Brockman et~al.(2016)Brockman, Cheung, Pettersson, Schneider,
  Schulman, Tang, and Zaremba]{gym}
Brockman, G., Cheung, V., Pettersson, L., Schneider, J., Schulman, J., Tang,
  J., and Zaremba, W.
\newblock {OpenAI Gym}.
\newblock \emph{arXiv preprint arXiv:1606.01540}, 2016.

\bibitem[Byrne \& Russon(1998)Byrne and Russon]{hil3}
Byrne, R.~W. and Russon, A.~E.
\newblock Learning by imitation: A hierarchical approach.
\newblock \emph{Behavioral and brain sciences}, 21\penalty0 (5):\penalty0
  667--684, 1998.

\bibitem[Daniel et~al.(2016)Daniel, Van~Hoof, Peters, and Neumann]{ML16}
Daniel, C., Van~Hoof, H., Peters, J., and Neumann, G.
\newblock Probabilistic inference for determining options in reinforcement
  learning.
\newblock \emph{Machine Learning}, 104\penalty0 (2-3):\penalty0 337--357, 2016.

\bibitem[Fei et~al.(2020)Fei, Wang, Zhuang, Zhang, Hao, Zhang, Ji, and
  Liu]{tripleGAIL}
Fei, C., Wang, B., Zhuang, Y., Zhang, Z., Hao, J., Zhang, H., Ji, X., and Liu,
  W.
\newblock {Triple-GAIL}: A multi-modal imitation learning framework with
  generative adversarial nets.
\newblock \emph{Proc. Int. Joint Conf. Artificial Intell.}, 2020.

\bibitem[Ghasemipour et~al.(2020)Ghasemipour, Zemel, and Gu]{fgail}
Ghasemipour, S. K.~S., Zemel, R., and Gu, S.
\newblock A divergence minimization perspective on imitation learning methods.
\newblock In \emph{Proc. Conf. Robot. Learn.}, 2020.

\bibitem[Goodfellow et~al.(2014)Goodfellow, Pouget-Abadie, Mirza, Xu,
  Warde-Farley, Ozair, Courville, and Bengio]{gan}
Goodfellow, I., Pouget-Abadie, J., Mirza, M., Xu, B., Warde-Farley, D., Ozair,
  S., Courville, A., and Bengio, Y.
\newblock Generative adversarial nets.
\newblock \emph{Proc. Advances in Neural Inf. Process. Syst.}, 2014.

\bibitem[Hamidi et~al.(2015)Hamidi, Tadepalli, Goetschalckx, and Fern]{hil5}
Hamidi, M., Tadepalli, P., Goetschalckx, R., and Fern, A.
\newblock Active imitation learning of hierarchical policies.
\newblock In \emph{Proc. Int. Joint Conf. Artificial Intell.}, 2015.

\bibitem[Henderson et~al.(2018)Henderson, Chang, Bacon, Meger, Pineau, and
  Precup]{optionGAN}
Henderson, P., Chang, W.-D., Bacon, P.-L., Meger, D., Pineau, J., and Precup,
  D.
\newblock {OptionGAN}: Learning joint reward-policy options using generative
  adversarial inverse reinforcement learning.
\newblock In \emph{Proc. {AAAI} Conf. Artificial Intell.}, 2018.

\bibitem[Ho \& Ermon(2016)Ho and Ermon]{GAIL}
Ho, J. and Ermon, S.
\newblock Generative adversarial imitation learning.
\newblock In \emph{Proc. Advances in Neural Inf. Process. Syst.}, 2016.

\bibitem[James et~al.(2020)James, Ma, Arrojo, and Davison]{rlbench}
James, S., Ma, Z., Arrojo, D.~R., and Davison, A.~J.
\newblock {RLBench}: The robot learning benchmark \& learning environment.
\newblock \emph{{IEEE} Robot. Auto. Letters}, 5\penalty0 (2):\penalty0
  3019--3026, 2020.

\bibitem[Kipf et~al.(2019)Kipf, Li, Dai, Zambaldi, Sanchez-Gonzalez,
  Grefenstette, Kohli, and Battaglia]{Compile}
Kipf, T., Li, Y., Dai, H., Zambaldi, V., Sanchez-Gonzalez, A., Grefenstette,
  E., Kohli, P., and Battaglia, P.
\newblock {CompILE}: Compositional imitation learning and execution.
\newblock In \emph{Proc. Int. Conf. Mach. Learn.}, 2019.

\bibitem[Le et~al.(2018)Le, Jiang, Agarwal, Dudik, Yue, and Daum{\'e}]{h-bc}
Le, H., Jiang, N., Agarwal, A., Dudik, M., Yue, Y., and Daum{\'e}, H.
\newblock Hierarchical imitation and reinforcement learning.
\newblock In \emph{Proc. Int. Conf. Mach. Learn.}, 2018.

\bibitem[Lee \& Seo(2020)Lee and Seo]{d-info-gail2}
Lee, S.-H. and Seo, S.-W.
\newblock Learning compound tasks without task-specific knowledge via imitation
  and self-supervised learning.
\newblock In \emph{Proc. Int. Conf. Mach. Learn.}, 2020.

\bibitem[Li et~al.(2021)Li, Song, and Tao]{Liskill}
Li, C., Song, D., and Tao, D.
\newblock The skill-action architecture: Learning abstract action embeddings
  for reinforcement learning.
\newblock \emph{submissions of ICLR}, 2021.

\bibitem[Maaten \& Hinton(2008)Maaten and Hinton]{tsne}
Maaten, L. v.~d. and Hinton, G.
\newblock Visualizing data using t-sne.
\newblock \emph{J. Mach. Learn. Res.}, 9\penalty0 (Nov):\penalty0 2579--2605,
  2008.

\bibitem[Nachum et~al.(2018)Nachum, Gu, Lee, and Levine]{hiro}
Nachum, O., Gu, S.~S., Lee, H., and Levine, S.
\newblock Data-efficient hierarchical reinforcement learning.
\newblock In \emph{Proc. Advances in Neural Inf. Process. Syst.}, 2018.

\bibitem[Pomerleau(1988)]{bc1}
Pomerleau, D.~A.
\newblock {ALVINN}: An autonomous land vehicle in a neural network.
\newblock In \emph{Proc. Advances in Neural Inf. Process. Syst.}, 1988.

\bibitem[Ross et~al.(2011)Ross, Gordon, and Bagnell]{compounding-error}
Ross, S., Gordon, G., and Bagnell, D.
\newblock A reduction of imitation learning and structured prediction to
  no-regret online learning.
\newblock In \emph{Proc. Int. Conf. Artificial Intell. \& Stat.}, 2011.

\bibitem[Schulman et~al.(2017)Schulman, Wolski, Dhariwal, Radford, and
  Klimov]{PPO}
Schulman, J., Wolski, F., Dhariwal, P., Radford, A., and Klimov, O.
\newblock Proximal policy optimization algorithms.
\newblock \emph{arXiv preprint arXiv:1707.06347}, 2017.

\bibitem[Sharma et~al.(2018)Sharma, Sharma, Rhinehart, and Kitani]{d-info-gail}
Sharma, M., Sharma, A., Rhinehart, N., and Kitani, K.~M.
\newblock {Directed-Info GAIL}: Learning hierarchical policies from unsegmented
  demonstrations using directed information.
\newblock In \emph{Proc. Int. Conf. Learn. Representations}, 2018.

\bibitem[Sharma et~al.(2019)Sharma, Pathak, and Gupta]{hil4}
Sharma, P., Pathak, D., and Gupta, A.
\newblock Third-person visual imitation learning via decoupled hierarchical
  controller.
\newblock In \emph{Proc. Advances in Neural Inf. Process. Syst.}, 2019.

\bibitem[Sutton et~al.(1999)Sutton, Precup, and Singh]{SMDP}
Sutton, R.~S., Precup, D., and Singh, S.
\newblock Between mdps and semi-mdps: A framework for temporal abstraction in
  reinforcement learning.
\newblock \emph{Artificial intelligence}, 112\penalty0 (1-2):\penalty0
  181--211, 1999.

\bibitem[Syed et~al.(2008)Syed, Bowling, and Schapire]{ocm}
Syed, U., Bowling, M., and Schapire, R.~E.
\newblock Apprenticeship learning using linear programming.
\newblock In \emph{Proc. Int. Conf. Mach. Learn.}, 2008.

\bibitem[Todorov et~al.(2012)Todorov, Erez, and Tassa]{mujoco}
Todorov, E., Erez, T., and Tassa, Y.
\newblock Mujoco: A physics engine for model-based control.
\newblock In \emph{Proc. {IEEE/RSJ} Int. Conf. Intelligent Robots \& Systems},
  2012.

\bibitem[Viterbi(1967)]{viterbi}
Viterbi, A.
\newblock Error bounds for convolutional codes and an asymptotically optimum
  decoding algorithm.
\newblock \emph{IEEE transactions on Information Theory}, 13\penalty0
  (2):\penalty0 260--269, 1967.

\bibitem[Yu et~al.(2018)Yu, Abbeel, Levine, and Finn]{hil2}
Yu, T., Abbeel, P., Levine, S., and Finn, C.
\newblock One-shot hierarchical imitation learning of compound visuomotor
  tasks.
\newblock \emph{arXiv preprint arXiv:1810.11043}, 2018.

\bibitem[Zhang \& Whiteson(2019)Zhang and Whiteson]{DAC}
Zhang, S. and Whiteson, S.
\newblock {DAC}: The double actor-critic architecture for learning options.
\newblock In \emph{Proc. Advances in Neural Inf. Process. Syst.}, 2019.

\bibitem[Zhang \& Paschalidis(2020)Zhang and Paschalidis]{provable_h-bc}
Zhang, Z. and Paschalidis, I.
\newblock Provable hierarchical imitation learning via em.
\newblock \emph{arXiv preprint arXiv:2010.03133}, 2020.

\end{thebibliography}
\bibliographystyle{icml2021}

\onecolumn
\appendix
\section{Appendix}
\subsection{Proofs for Propositions and Theorems}
    Here we provide the proofs for the propositions and theorems used in our \method.

\subsubsection{Proof for $\gO_\text{one-step} = \gO$}
    According to Assumption 1.1, an option can be activated at any state, thus the intra-option policy $\pi_o(a|s)$, break policy $\beta_{o'}(s)$ and inter-option policy $\pi_\gO(o|s)$ are all well-defined on any $s \in \sS, \forall o \in \gO$. This suggests that $\pi_L(a|s, o) \equiv \pi_o(a|s)$ holds over all options on any state. For $\beta_{o'}(s)$, with Assumption 1.2, we have $\beta_{o'}(s) = 1-\pi_H(o'|s, o')$ and for $\pi_\gO (o|s)$ we have $\pi_\gO (o|s) = \left.\frac{\pi_H(o|s, o')}{\sum_{o \neq o'} \pi_H(o|s, o')}\right|_{\forall o' \neq o} = \frac{\sum_{o' \ne o}\pi_H(o|s, o')}{\sum_{o' \ne o}\sum_{o \neq o'} \pi_H(o|s, o')}$. Also, with $o_{-1} \equiv \#$, it can be directly found that $\tilde{\mu}_0(s, o) = \tilde{\mu}_0(s, o=\#) \equiv \mu_0(s)$.
    Since $\sS, \sA, R_s^a, P_{s, s'}^a, \gamma$ are all defined the same between $\gO_\text{one-step}$ and $\gO$, we can get that$\gO_\text{one-step} = \gO$ holds under Assumption 1, and there exists an one-to-one mapping between $\left(\pi_H(o|s, o'), \pi_L(a|s, o)\right)$ and $\left(\pi_o(a|s), \beta_{o'}(s), \pi_\gO(o|s)\right)$.\hfill$\square\qquad\qquad\qquad\qquad$
    
    Combining with Theorem 1, this equivalency also suggests:
    \begin{gather}
        \rho_{\tilde{\pi}}(s, a, o, o') = \rho_{\tilde{\pi}^\star}(s, a, o, o') \Leftrightarrow \tilde{\pi} = \tilde{\pi}^\star \Leftrightarrow \left(\pi_o(a|s), \pi_\gO(o|s), \beta_{o'}(s)\right) = \left(\pi_o^\star(a|s), \pi_\gO^\star(o|s), \beta_{o'}^\star(s)\right).
    \end{gather}
    
\subsubsection{Proof for Theorem 1}
    The proof of Theorem 1 can be derived similar as that from \citet{ocm} by defining an augmented MDP with options: $\tilde{s}_t \doteq (s_t, o_{t-1}) \in \sS \times \sO^+, \tilde{a}_t \doteq (a_t, o^A_t) \in \sA \times \sO, \tilde{\pi}(\tilde{a}_t| \tilde{s}_t) \doteq \pi_L(a_t|s_t, o^A_t) \pi_H(o^A_t|s_t, o_{t-1}), \tilde{P}_{\tilde{s}_t, \tilde{s}_{t+1}}^{\tilde{a}_t} \doteq P_{s_t, s_{t+1}}^{a_t} \mathds{1}_{o_t = o^A_t}$, where we denote $o_t$ used in $\tilde{a}_t$ as $o^A_t$ for better distinguish from the option chosen in $\tilde{s}_{t+1}$, despite they should actually be the same.
    \begin{figure}[h!] \begin{center}
        \includegraphics[width=0.3\linewidth]{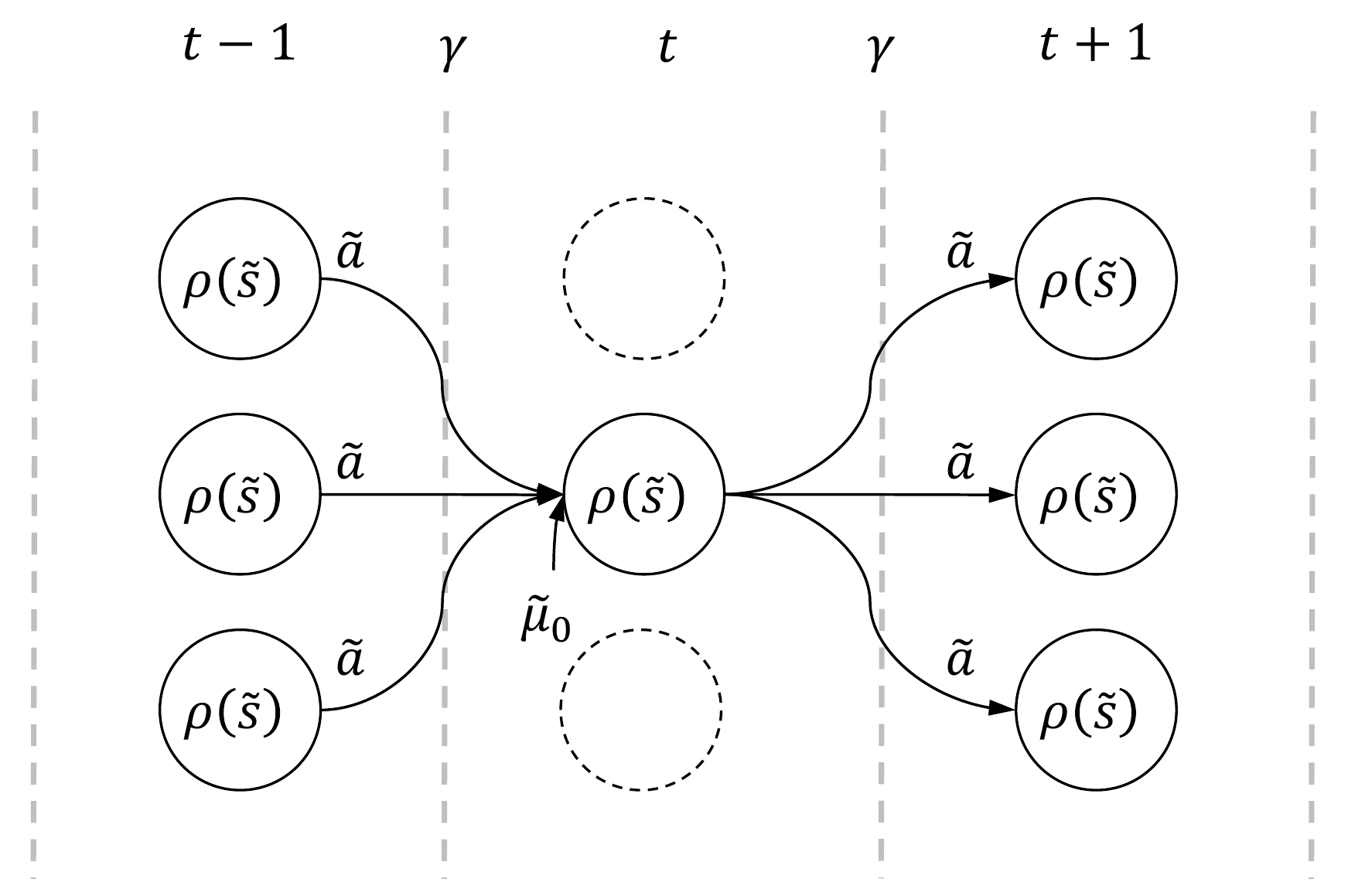}
        \caption{Illustration of the Bellman Flow on augmented MDP with options.}
        \label{fig:bellman_flow}
    \end{center} \end{figure}
    
    With the sugmented MDP, we can rewrite:
    \begin{eqnarray}
        \rho(\tilde{s}, \tilde{a}) &\doteq& \rho(s, a, o, o') \nonumber\\
       &=& \pi_L(a|s, o) \pi_L(o|s, o') \left(\tilde{\mu}_0(s, o') + \gamma \sum_{s', a', o''} \rho(s', a', o', o'') P_{s', s}^{a'}\right) \\
       &=& \tilde{\pi}(\tilde{a}|\tilde{s})\left(\tilde{\mu}_0 + \gamma \sum_{\tilde{s}', \tilde{a}'}\rho(\tilde{s}', \tilde{a}')\tilde{P}_{\tilde{s}', \tilde{s}}^{\tilde{a}'}\right)\nonumber
    \end{eqnarray}
    and construct a $\tilde{\pi}$-specific Bellman Flow constraint similar as that introduced by \citet{ocm}:
    \begin{eqnarray}
        \rho(\tilde{s}, \tilde{a}) &=& \tilde{\pi}(\tilde{a}|\tilde{s}) \left(\tilde{\mu}_0(\tilde{s}) + \gamma \sum_{\tilde{s}', \tilde{a}'} \rho(\tilde{s}', \tilde{a}')\tilde{P}_{\tilde{s}', \tilde{s}}^{\tilde{a}'}\right)\label{eq:ocm_bellman_flow}\\
        \rho(\tilde{s}, \tilde{a}) &\geq& 0.\label{eq:ocm_non_neg}
    \end{eqnarray}
    
    Now we build the relation between the option-occupancy measurement $\rho_{\tilde{\pi}}(\tilde{s}, \tilde{a})$ and the policy $\tilde{\pi}(\tilde{a}| \tilde{s})$.
    
    \begin{lemma}\label{lam:solution}
        The option-occupancy measurement of $\tilde{\pi}$ which is defined as $\rho_{\tilde{\pi}}(\tilde{s}, \tilde{a}) = \mathbb{E}\left[\sum_{t=0}^\infty \gamma^t \mathds{1}_{(\tilde{s}_t=\tilde{s},\tilde{a}_t=\tilde{a})}\right]$ satisfies the $\tilde{\pi}$-specific Bellman Flow constraint in \Eqref{eq:ocm_bellman_flow}-\ref{eq:ocm_non_neg}.
    \end{lemma}
    
    \textit{proof:} it can be directly find that \Eqref{eq:ocm_non_neg} is always satisfied as $\rho_{\tilde{\pi}}(\tilde{s}, \tilde{a}) = \mathbb{E}\left[\sum_{t=0}^\infty \gamma^t \mathds{1}_{(\tilde{s}_t=\tilde{s},\tilde{a}_t=\tilde{a})}\right] \geq 0$ always holds, we now verify the constraint in \Eqref{eq:ocm_bellman_flow}:
    \begin{eqnarray}
        \rho_{\tilde{\pi}}(\tilde{s},\tilde{a}) &=& \mathbb{E}\left[\sum_{t=0}^\infty \gamma^t \mathds{1}_{(\tilde{s}_t=\tilde{s},\tilde{a}_t=\tilde{a})}\right]
        = \sum_{t=0}^\infty \gamma^t P(\tilde{s}_t=\tilde{s}, \tilde{a}_t=\tilde{a})\\
       &=& \tilde{\pi}(\tilde{a}|\tilde{s}) \tilde{\mu}_0(\tilde{s}) + \sum_{t=1}^\infty \gamma^t P(\tilde{s}_t = \tilde{s}, \tilde{a}_t = \tilde{a})\\
       &=& \tilde{\pi} (\tilde{a}|\tilde{s}) \tilde{\mu}_0(\tilde{s}) + \sum_{t=1}^\infty \gamma^t \sum_{\tilde{s}', \tilde{a}'} P(\tilde{s}_t = \tilde{s}, \tilde{a}_t = \tilde{a}, \tilde{s}_{t-1} = \tilde{s}', \tilde{a}_{t-1} = \tilde{a}')\\
       &=& \tilde{\pi}(\tilde{a}|\tilde{s}) \left(\tilde{\mu}_0(\tilde{s}) + \sum_{t=1}^\infty \gamma^t \sum_{\tilde{s}', \tilde{a}'} \tilde{P}_{\tilde{s}', \tilde{s}}^{\tilde{a}'} P(\tilde{s}_{t-1} = \tilde{s}', \tilde{a}_{t-1} = \tilde{a}')\right) \\
       &=& \tilde{\pi}(\tilde{a}|\tilde{s}) \left(\tilde{\mu}_0(\tilde{s}) + \gamma \sum_{\tilde{s}', \tilde{a}'} \tilde{P}_{\tilde{s}', \tilde{s}}^{\tilde{a}'} \sum_{t=0}^\infty \gamma^t  P(\tilde{s}_t = \tilde{s}', \tilde{a}_t = \tilde{a}')\right) \\
       &=& \tilde{\pi}(\tilde{a}|\tilde{s}) \left(\tilde{\mu}_0(\tilde{s}) + \gamma \sum_{\tilde{s}', \tilde{a}'} \tilde{P}_{\tilde{s}', \tilde{s}}^{\tilde{a}'} \mathbb{E}_{\tilde{\pi}} \left[\sum_{t=0}^\infty \gamma^t \mathds{1}_{(\tilde{s}_t = \tilde{s}', \tilde{a}_t = \tilde{a}')}\right]\right)\\
       &=& \tilde{\pi}(\tilde{a}|\tilde{s}) \left( \tilde{\mu}_0(\tilde{s}) + \gamma \sum_{\tilde{s}', \tilde{a}'} \rho_{\tilde{\pi}} (\tilde{s}',\tilde{a}') \tilde{P}_{\tilde{s}', \tilde{s}}^{\tilde{a}'} \right)\qquad\qquad\qquad\qquad\square
    \end{eqnarray}

    \begin{lemma}\label{lam:unique}
        The function that satisfies the $\tilde{\pi}$-specific Bellman Flow constraint in \Eqref{eq:ocm_bellman_flow}-\ref{eq:ocm_non_neg} is unique.
    \end{lemma}
    \textit{proof:} we first define an operator for policy $\tilde{\pi}$: $\gT^{\tilde{\pi}}: R^{|\sS \times \sO^+|\times|\sA \times \sO|} \mapsto R^{|\sS \times \sO^+|\times|\sA \times \sO|}$ for any function $f \in R^{|\sS \times \sO^+|\times|\sA \times \sO|}$: $\left(\gT^{\tilde{\pi}}f\right)(\tilde{s}, \tilde{a}) \doteq \tilde{\pi}(\tilde{a}|\tilde{s})\left(\tilde{\mu}_0(\tilde{s}) + \gamma \sum_{\tilde{s}', \tilde{a}'} f(\tilde{s}', \tilde{a}') \tilde{P}_{\tilde{s}', \tilde{s}}^{\tilde{a}'}\right) $, then for any two functions $ \rho_{1}(\tilde{s}, \tilde{a}) \geq 0, \rho_{2}(\tilde{s}, \tilde{a}) \geq 0$ satisfy $ \rho_{1} = \gT^{\tilde{\pi}}\rho_{1}, \rho_{2} = \gT^{\tilde{\pi}}\rho_{2}$, we have:
    \begin{eqnarray}
         &&\sum_{\tilde{s}, \tilde{a}} \left|\rho_{1} - \rho_{2}\right|(\tilde{s}, \tilde{a}) = \sum_{\tilde{s}, \tilde{a}}  \left|\gT^{\tilde{\pi}}\rho_{1} - \gT^{\tilde{\pi}}\rho_{2}\right|(\tilde{s}', \tilde{a}')\\
        &=& \sum_{\tilde{s}, \tilde{a}} \left| \tilde{\pi}(\tilde{a}|\tilde{s}) \gamma \sum_{\tilde{s}', \tilde{a}'} \tilde{P}_{\tilde{s}', \tilde{s}}^{\tilde{a}'} \left(\rho_{1}-\rho_{2}\right)(\tilde{s}', \tilde{a}')\right| = \gamma \sum_{\tilde{s}, \tilde{a}} \left| \sum_{\tilde{s}', \tilde{a}'} p(\tilde{s}, \tilde{a}|\tilde{s}', \tilde{a}') \left(\rho_{1}-\rho_{2}\right)(\tilde{s}', \tilde{a}')\right|\\
        &\leq &\gamma \sum_{\tilde{s}, \tilde{a}} \sum_{\tilde{s}', \tilde{a}'} p(\tilde{s}, \tilde{a}|\tilde{s}', \tilde{a}') \left|\rho_{1}-\rho_{2}\right|(\tilde{s}', \tilde{a}') = \gamma \sum_{\tilde{s}', \tilde{a}'}\left|\rho_{1} - \rho_{2}\right|(\tilde{s}', \tilde{a}')\\
        &=& \gamma \sum_{\tilde{s}, \tilde{a}}\left|\rho_{1} - \rho_{2}\right|(\tilde{s}, \tilde{a})\\
        &\because & \sum_{\tilde{s}, \tilde{a}} \left|\rho_{1} - \rho_{2}\right|(\tilde{s}, \tilde{a}) \geq 0, \gamma < 1\\
        &\therefore &\sum_{\tilde{s}, \tilde{a}}\left|\rho_{1} - \rho_{2}\right|(\tilde{s}, \tilde{a}) = 0 \Rightarrow  \rho_{1} = \rho_{2} \qquad\qquad\qquad\qquad\qquad\qquad\qquad\qquad\square
    \end{eqnarray}
    
    \begin{lemma}\label{lam:policy}
        There is a bijection between $\tilde{\pi}(\tilde{a}|\tilde{s})$ and $\left(\pi_H(o|s, o'), \pi_L(a|s, o)\right)$, where $\tilde{\pi}(\tilde{a}|\tilde{s}) = \tilde{\pi}(a, o|s, o') = \pi_L(a|s, o) \pi_H(o|s, o')$ and $\pi_H(o|s, o') = \sum_{a}\tilde{\pi}(a, o|s, o'), \pi_L(a|s, o) = \left.\frac{\tilde{\pi}(a, o|s, o')}{\sum_{a}\tilde{\pi}(a, o|s, o')}\right|_{\forall o'} = \frac{\sum_{o'}\tilde{\pi}(a, o|s, o')}{\sum_{a, o'}\tilde{\pi}(a, o|s, o')}$
    \end{lemma}

    With Lemma~\ref{lam:solution} and Lemma~\ref{lam:unique}, the proof of Theorem 1 is provided:
    
    \textit{proof:} For any $\rho(s, a, o, o') = \rho(\tilde{s}, \tilde{a}) \in \sD = \left\{\rho(\tilde{s}, \tilde{a}) \geq 0; \sum_{\tilde{a}} \rho(\tilde{s}, \tilde{a})=\tilde{\mu}_0(\tilde{s})+\gamma \sum_{\tilde{s}', \tilde{a}'} \rho(\tilde{s}',\tilde{a}')\tilde{P}_{\tilde{s}', \tilde{s}}^{\tilde{a}'}\right\}$, and a policy $\tilde{\pi}(\tilde{a}|\tilde{s})$ satisfies:
    \begin{align}
        \tilde{\pi}(\tilde{a}|\tilde{s}) = \frac{\rho(\tilde{s}, \tilde{a})}{\sum_{\tilde{a}}\rho(\tilde{s}, \tilde{a})} = \frac{\rho(\tilde{s}, \tilde{a})}{\tilde{\mu}_0(\tilde{s})+\gamma \sum_{\tilde{s}', \tilde{a}'} \rho(\tilde{s}',\tilde{a}')\tilde{P}_{\tilde{s}', \tilde{s}}^{\tilde{a}'}},\label{eq:ocm_f1}.
    \end{align}
    With \Eqref{eq:ocm_f1} $\rho$ should be a solution of \Eqref{eq:ocm_bellman_flow}-\ref{eq:ocm_non_neg}, and with Lemma~\ref{lam:solution}-\ref{lam:unique}, the solution is unique and equals to the occupancy measurement of $\tilde{\pi}$. With Lemma~\ref{lam:policy}, $\rho$ is also the unique occupancy measurement of $(\pi_H, \pi_L)$.
    
    On the other hand, If $\rho_{\tilde{\pi}}$ is the occupancy measurement of $\tilde{\pi}$, we have:
    \begin{align}
        \sum_{\tilde{a}} \tilde{\pi}(\tilde{a}|\tilde{s}) = 1 = \frac{\sum_{\tilde{a}}{\rho_{\tilde{\pi}}(\tilde{s}, \tilde{a})}}{\tilde{\mu}_0(\tilde{s})+\gamma \sum_{\tilde{s}', \tilde{a}'}\rho_{\tilde{\pi}}(\tilde{s}',\tilde{a}')\tilde{P}_{\tilde{s}', \tilde{s}}^{\tilde{a}'}},
    \end{align}
    which indicates that $\rho_{\tilde{\pi}} \in \sD$ and 
    $\tilde{\pi}(a, o|s, o') = \frac{\rho_{\tilde{\pi}}(s, a, o, o')}{\sum_{a, o}\rho_{\tilde{\pi}}(s, a, o, o')}$, also: 
    \begin{align}
        \pi_H(o|s, o') &= \sum_{a}\tilde{\pi}(a, o|s, o') = \frac{\sum_{a}\rho_{\tilde{\pi}}(s, a, o, o')}{\sum_{a, o}\rho_{\tilde{\pi}}(s, a, o, o')}\\
        \pi_L(a|s, o) &= \frac{\sum_{o'}\tilde{\pi}(a, o|s, o')}{\sum_{a, o'}\tilde{\pi}(a, o|s, o')} = \frac{\sum{o'}\rho_{\tilde{\pi}}(s, a, o, o')}{\sum_{a, o'}\rho_{\tilde{\pi}}(s, a, o, o')} &\square
    \end{align}
    
\subsubsection{Proof for Theorem 2}
    We first adapt the corollary on \citet{fgail} into its option-version.
    \begin{lemma}\label{lam:rlirl}
        Optimizing the $f$-divergence between $\rho_{\tilde{\pi}}$ and $\rho_{\tilde{\pi}_E}$ equals to perform $\tilde{\pi}^\star = \text{HRL}(c^\star)$ with $c^\star = \text{HIRL}_{\psi}(\tilde{\pi}_E)$: $\tilde{\pi}^\star = \text{HRL}\circ\text{HIRL}_{\psi}(\tilde{\pi}_E) = \argmin_{\tilde{\pi}} -\mathbb{H}(\tilde{\pi}) + D_f\left(\rho_{\tilde{\pi}}(s, a, o, o')\|\rho_{\tilde{\pi}_E}(s, a, o, o')\right)$
    \end{lemma}
     \textit{proof:} we take similar deviations from that provided by \citet{fgail}. Let $f$ be a function defining a $f$-divergence and let $f^\star$ be the convex conjugate of $f$. Given $\rho_{\tilde{\pi}_E}$ and cost functions $c(s, a, o, o')$ defined on $\sS \times \sA \times \sO \times \sO^+$, we can define the cost function regularizer used by our option-based HIRL as $\psi_f(c) \doteq \mathbb{E}_{\rho_{\tilde{\pi}_E}(s, a, o, o')}\left[f^\star\left(c(s, a, o, o')\right) - c(s, a, o, o')\right]$ and a similar relation holds:
    \begin{eqnarray}
         &&\psi_{f}^\star\left(\rho_{\tilde{\pi}}(s, a, o, o') - \rho_{\tilde{\pi}_E}(s, a, o, o')\right)\\
        &=&\sup_{c(\dot)} \left[\sum_{s, a, o, o'}(\rho_{\tilde{\pi}} - \rho_{\tilde{\pi}_E})(s, a, o, o')c(s, a, o, o') - \psi_f(c)\right]\\
        &=&\sup_{c(\dot)} \left[\sum_{s, a, o, o'}(\rho_{\tilde{\pi}} - \rho_{\tilde{\pi}_E})(s, a, o, o')c(s, a, o, o')\right.\nonumber\\
         &&\qquad\qquad\qquad\left.- \sum_{s, a, o, o'} \rho_{\tilde{\pi}_E}(s, a, o, o') \left(f^\star\left(c(s, a, o, o')\right) - c(s, a, o, o')\right)\right]\\
        &=&\sup_{c(\dot)} \left[\sum_{s, a, o, o'}\left[\rho_{\tilde{\pi}}(s, a, o, o')c(s, a, o, o') - \rho_{\tilde{\pi}_E}(s, a, o, o')f^\star\left( c(s, a, o, o')\right)\right]\right]\\
        &=&\sup_{c(\dot)} \left[\mathbb{E}_{\rho_{\tilde{\pi}}}\left[c(s, a, o, o')\right] - \mathbb{E}_{\rho_{\tilde{\pi}_E}}\left[f^\star\left( c(s, a, o, o')\right)\right]\right], \text{  let  } T_{\omega} = c\\
        &=&\sup_{T_{\omega}} \left[\mathbb{E}_{\rho_{\tilde{\pi}}}\left[T_{\omega}(s, a, o, o')\right] - \mathbb{E}_{\rho_{\tilde{\pi}_E}}\left[f^\star\left( T_{\omega}(s, a, o, o')\right)\right]\right]\\
        &=&D_f\left(\rho_{\tilde{\pi}}(s, a, o, o')\|\rho_{\tilde{\pi}_E}(s, a, o, o')\right),
    \end{eqnarray}
    where $\tilde{\pi}^\star = \text{HRL}\circ\text{HIRL}_{\psi}(\tilde{\pi}_E) = \argmin_{\tilde{\pi}} - \mathbb{H}(\tilde{\pi}) + \psi_{f}^\star\left(\rho_{\tilde{\pi}}(s, a, o, o') - \rho_{\tilde{\pi}_E}(s, a, o, o')\right) = \argmin_{\tilde{\pi}} - \mathbb{H}(\tilde{\pi}) + D_f\left(\rho_{\tilde{\pi}}(s, a, o, o')\|\rho_{\tilde{\pi}_E}(s, a, o, o')\right)$. \hfill$\square\qquad\qquad\qquad\qquad$

    Similar as \citet{fgail}, we omit the entropy regularizer term in Lemma~\ref{lam:rlirl}, thus after the optimization in M-step we have
    $D_f\left(\rho_{\tilde{\pi}^{n-1}}(s, a, o, o')\|\rho_{E}(s, a) [p_{\tilde{\pi}^n}(o, o'| s, a)\right) \geq D_f\left(\rho_{\tilde{\pi}^{n}}(s, a, o, o')\|\rho_{E}(s, a)p_{\tilde{\pi}^n}(o, o'| s, a)\right)$. Now we are ready for proving Theorem 2:
    
    \textit{proof:} Since the option of expert is inferred based on the policy $\tilde{\pi}^n$ on each optimization step, we separate the expert option-occupancy measurement estimated with $\tilde{\pi}^n$ as: $\rho_{\tilde{\pi}_E}(s, a, o, o') = \rho_{E}(s, a) p_{\tilde{\pi}^{n}}(o,o'|s,a)$.
    By repeating the definition of $Q_n$ in our main paper, we have
    \begin{eqnarray}
        Q_n &=& \mathbb{E}_{p_{\tilde{\pi}^{n-1}}(o,o'|s,a)}\left[D_f\left(\rho_{\tilde{\pi}^{n}}(s, a, o, o')\|\rho_{\tilde{\pi}_E}(s, a, o, o')\right)\right] \label{eq:th2:saoo1}\\
        &=& \sum_{s, a, o, o'} \rho_{E}(s, a) p_{\tilde{\pi}^{n-1}}(o,o'|s,a) f\left(\frac{\rho_{\tilde{\pi}^n}(s, a, o, o')}{\rho_{E}(s, a) p_{\tilde{\pi}^{n-1}}(o,o'|s,a)}\right) \nonumber\\
        &\geq&
        \sum_{s, a} \rho_{E}(s, a) f\left(\frac{\rho_{\tilde{\pi}^n}(s, a)}{\rho_{E}(s, a)}\right) \quad(f \text{ is convex}) \label{eq:th2:sa}\\
        &=& \sum_{s, a, o, o'} \rho_{E}(s, a) p_{\tilde{\pi}^n}(o,o'|s,a) f\left(\frac{\rho_{\tilde{\pi}^n}(s, a, o, o')}{\rho_{E}(s, a) p_{\tilde{\pi}^n}(o,o'|s,a)}\right) \quad(\text{E-Step})\nonumber\\
        &\geq& \sum_{s, a, o, o'} \rho_{E}(s, a) p_{\tilde{\pi}^n}(o,o'|s,a) f\left(\frac{\rho_{\tilde{\pi}^{n+1}}(s, a, o, o')}{\rho_{E}(s, a) p_{\tilde{\pi}^n}(o,o'|s,a)}\right) \quad(\text{M-Step}) \nonumber\\
        &=& Q_{n+1}. \qquad\qquad\qquad\qquad\qquad\qquad\qquad\qquad\qquad\qquad\qquad\qquad\qquad\square\label{eq:th2:saoo2}
    \end{eqnarray}
    
    With \Eqref{eq:th2:saoo1}, \Eqref{eq:th2:sa} and \Eqref{eq:th2:saoo2} we can also obtain:
    \begin{eqnarray}
        D_f\left(\rho_{\tilde{\pi}^n}(s, a, o, o')\|\rho_{E}(s, a) p_{\tilde{\pi}^n}(o, o'|s, a)\right) &\geq& D_f\left(\rho_{\tilde{\pi}^{n+1}}(s, a, o, o')\|\rho_{E}(s, a) p_{\tilde{\pi}^{n+1}}(o, o'|s, a)\right)\\
        \Rightarrow\qquad\qquad\qquad\qquad D_f\left(\rho_{\tilde{\pi}^n}(s, a)\|\rho_{E}(s, a)\right) &\geq& D_f\left(\rho_{\tilde{\pi}^{n+1}}(s, a)\|\rho_{E}(s, a)\right)
    \end{eqnarray}
    
\subsection{Experimental Details and Extra Results}
    Here we provide more comparative results on several counterparts, as well as the experimental details.\footnote{The source code is provided at \href{ https://github.com/id9502/Option-GAIL.git}{Option-GAIL.git}. For setting up the environments correctly, please also refer to \href{https://gym.openai.com/}{OpenAI-Gym}~\citep{gym} and \href{https://sites.google.com/view/rlbench}{RLBench}~\citep{rlbench}}.
    
\begin{figure}[h] \begin{center}
    \includegraphics[width=0.95\linewidth]{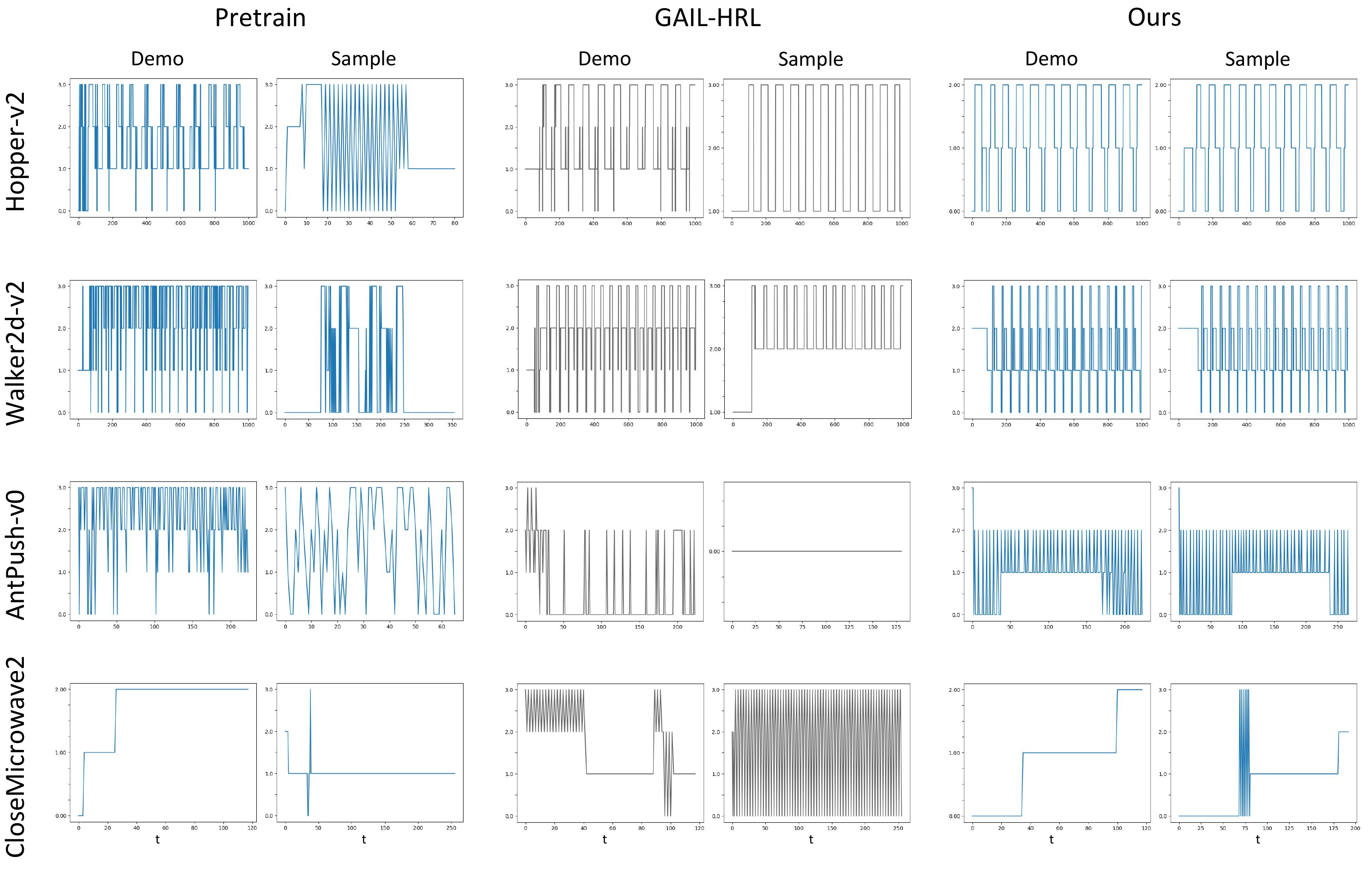}
    \caption{Visualization of the options activated at each step, learned respectively by pretraining and fixing high-level policy(Pretrain, refers to Directed-info GAIL~\citep{d-info-gail}), Mixer of Expert(MoE, refers to OptionGAN~\citep{optionGAN}), GAIL-HRL and our proposed method. 'Demo' denotes the options inferred from the expert, and 'Sample' denotes the options used by agent when doing self-explorations. The effectiveness of our proposed method on regularizing the option switching is obvious by comparing the consistent switching tendencies between Demo and Sample.}
    \label{fig:result_detail_opt}
\end{center} \end{figure}

\subsubsection{Extra Results}
\begin{figure}[h!] \begin{center}
    \includegraphics[width=0.95\linewidth]{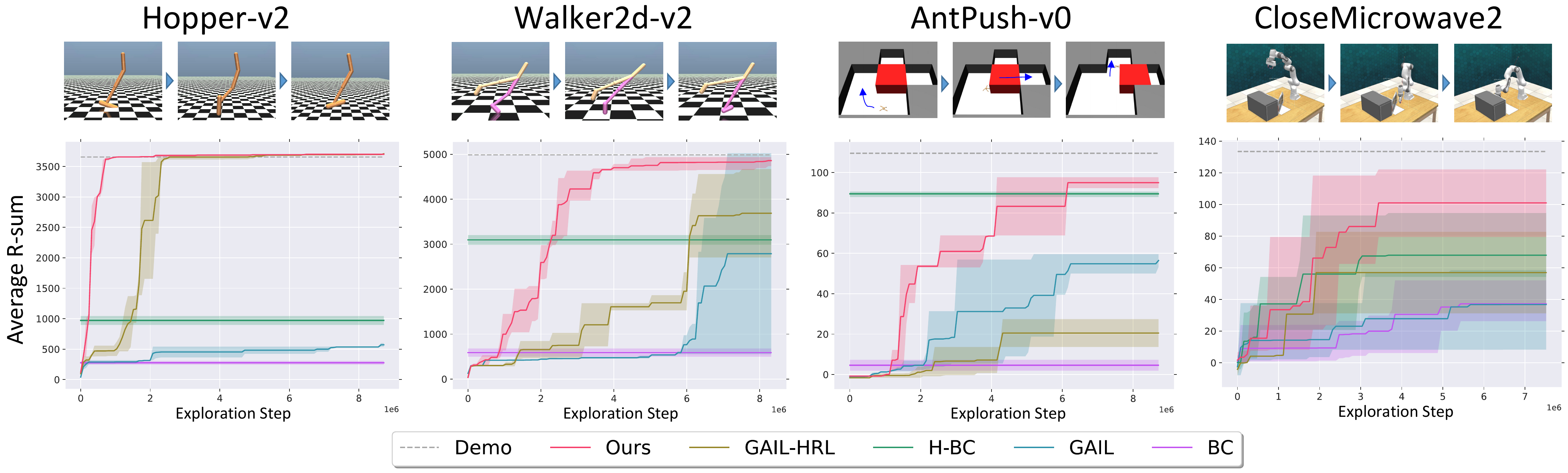}
    \caption{comparison of learning performance on four environments. We compare the \textbf{maximum average reward-sums} vs. exploration steps on different environments. The solid line indicates the average performance among several trials under different random seeds, while the shade indicates the range of the maximum average reward-sums over different trials.}
    \label{fig:result_compare}
\end{center} \end{figure}
\begin{table}[h!] \begin{center}
    \caption{\label{tab:comp_ex} Comparative results. All results are measured by the average \textbf{maximum average reward-sum} among different trails.}
    \begin{tabular}{ccccc}
    \toprule 
    & \small{Hopper-v2}& \small{Walker2d-v2} & \small{AntPush-v0} &\small{CloseMicrowave2} \\
    \midrule
    Demos $(s, a)\times T$ & $(\mathbb{R}^{11}, \mathbb{R}^3) \times 1$k & $(\mathbb{R}^{17}, \mathbb{R}^6) \times 5$k & $(\mathbb{R}^{107}, \mathbb{R}^8) \times 50$k & $(\mathbb{R}^{101}, \mathbb{R}^8) \times 1$k\\
    \small{Demo Reward} & 3656.17$\pm$0.0 & 5005.80$\pm$36.18 & 116.60$\pm$14.07 & --- \\
    \midrule
    \small{GAIL} & 535.29$\pm$7.19 & 2787.87$\pm$2234.46 & 56.45$\pm$3.17 & 39.14$\pm$12.87\\
    \small{Pretrain} & 436.55$\pm$27.74 & 891.70$\pm$100.58 & -0.07$\pm$1.50 & 74.34$\pm$20.16\\
    \small{MoE} & 3254.12$\pm$446.78 & 2722.11$\pm$2217.80 & 39.73$\pm$37.00 & 33.33$\pm$25.07\\
    \small{GAIL-HRL} & 3697.40$\pm$1.14 & 3687.63$\pm$982.99 & 20.53$\pm$6.90 & 56.95$\pm$25.74\\
    \small{Ours} & \textbf{3700.42$\pm$1.70} & \textbf{4836.85$\pm$100.09} & \textbf{95.00$\pm$2.70} & \textbf{100.74$\pm$21.33}\\
    \bottomrule 
    \end{tabular}
\end{center} \end{table}

\subsubsection{Experimental Details}

\begin{table}[h]
    \centering
    \begin{tabular}{c|cc}
        $|O|$ & Option-Viterbi / total (s) & $\%$ \\
    \midrule
        2 & $0.0938 / 57.785$ & $0.16\%$ \\
        3 & $0.0884 / 90.199$ & $0.10\%$ \\
        4 & $0.0840 / 102.00$ & $0.08\%$ \\
        5 & $0.0938 / 126.05$ & $0.07\%$ \\
        6 & $0.1014 / 142.64$ & $0.07\%$
    \end{tabular}
    \caption{The computation time of Option-Viterbi comparing with the overall learning time costs}
    \label{tab:time_costs}
\end{table}
\begin{table}[h!] \begin{center}
    \caption{\label{tab:config} Configurations and hyper-parameters}
    \begin{tabular}{c|c||c|c}
    \toprule 
    Name & Value & Name & Value \\
    \midrule
    $\gamma$ & 0.99 & learning rate & 0.0003\\
    $\lambda_{\sM_L}$ & 0 & $\lambda_{\sM^H}$ & 0.01\\
    batch size(T) & 4096 & mini batch size & 64\\
    \bottomrule 
    \end{tabular}
\end{center} \end{table}

\end{document}